\documentclass[twocolumn]{svjour3} 
\smartqed
\usepackage{natbib}
\usepackage{graphicx}

\usepackage{booktabs,multirow,array}
\usepackage{hyperref}
\usepackage{ragged2e}
\usepackage[ruled]{algorithm2e}
\usepackage{algorithmic}
\usepackage[dvipsnames]{xcolor}
\usepackage[misc]{ifsym}

\usepackage{amsmath}
\usepackage{algorithmic}
\usepackage[caption=false,font=footnotesize]{subfig}
\usepackage{ragged2e}

\begin{document}

\title{Bridging Composite and Real: Towards End-to-end Deep Image Matting

\thanks{*J. Li and J. Zhang are co-first authors and contribute equally to this work.\\This work was supported by Australian Research Council Projects  FL-170100117, IH-180100002, IC-190100031.}
}

\author{Jizhizi Li$^{1*}$,
        Jing Zhang$^{1*}$,
        Stephen J. Maybank$^{2}$,
        Dacheng Tao$^{1}$
    }

\institute{
Jizhizi Li (jili8515@uni.sydney.edu.au) \\
Jing Zhang (jing.zhang1@sydney.edu.au) \\
Stephen J. Maybank (sjmaybank@dcs.bbk.ac.uk) \\
\Letter~Dacheng Tao (dacheng.tao@sydney.edu.au)\\
$^{1}$ School of Computer Science,
Faculty of Engineering, The University of Sydney, Darlington,
NSW 2008, Australia.\\
$^{2}$ 
Department of Computer Science and Information System, Birkbeck College, University of London, U.K.
}

\date{Received: date / Accepted: date}
\maketitle

\begin{abstract}

Extracting accurate foregrounds from natural images benefits many downstream applications such as film production and augmented reality. However, the furry characteristics and various appearance of the foregrounds, e.g., animal and portrait, challenge existing matting methods, which usually require extra user inputs such as trimap or scribbles. To resolve these problems, we study the distinct roles of semantics and details for image matting and decompose the task into two parallel sub-tasks: high-level semantic segmentation and low-level details matting. Specifically, we propose a novel Glance and Focus Matting network (GFM), which employs a shared encoder and two separate decoders to learn both tasks in a collaborative manner for end-to-end natural image matting. Besides, due to the limitation of available natural images in the matting task, previous methods typically adopt composite images for training and evaluation, which result in limited generalization ability on real-world images. In this paper, we investigate the domain gap issue between composite images and real-world images systematically by conducting comprehensive analyses of various discrepancies between the foreground and background images. We find that a carefully designed composition route RSSN that aims to reduce the discrepancies can lead to a better model with remarkable generalization ability. Furthermore, we provide a benchmark containing 2,000 high-resolution real-world animal images and 10,000 portrait images along with their manually labeled alpha mattes to serve as a test bed for evaluating matting model's generalization ability on real-world images. Comprehensive empirical studies have demonstrated that GFM outperforms state-of-the-art methods and effectively reduces the generalization error. The code and the datasets will be released at \url{https://github.com/JizhiziLi/GFM}.


\end{abstract}

\section{Introduction}
\label{section: introduction}
Image matting refers to extracting the foreground alpha matte from an input image, requiring both hard labels for the explicit foreground or background and soft labels for the transition areas, which plays an important role in many applications, e.g., virtual reality, augmented reality, entertainment, etc. Typical foregrounds in image matting have furry details and diverse appearance, e.g., animal and portrait, which lay a great burden on image matting methods. How to recognize the semantic foregrounds or backgrounds as well as extracting the fine details for trimap-free natural image matting remains challenging in the image matting community.

For image matting, an image \textbf{I} is assumed to be a linear combination of foreground \textbf{F} and background \textbf{B} via a soft alpha matte $\alpha \in \left[0,1\right]$, $i.e.$,
\begin{equation}
\textbf{I}_i = \alpha_i \textbf{F}_i + \left(1-\alpha_i \right) \textbf{B}_i,
\label{equa:blending}
\end{equation}
where $i$ denotes the pixel index. It is a typical ill-posed problem to estimate \textbf{F}, \textbf{B}, and $\alpha$ given \textbf{I} from Eq.~\eqref{equa:blending} due to the under-determined nature. To relieve the burden, previous matting methods adopt extra user inputs such as trimaps~\citep{xu2017deep} and scribbles~\citep{levin2007closed} as priors to decrease the degree of unknown. Based on sampling neighboring known pixels~\citep{wang2007optimized,ruzon2000alpha,wang2005iterative} or defining an affinity matrix~\citep{zheng2008fuzzymatte}, the known alpha values ($i.e.$, foreground or background) are propagated to the unknown pixels. Usually, some edge-aware smoothness constraints are used to make the problem tractable~\citep{levin2007closed}. However, either the sampling or calculating affinity matrix is based on low-level color or structural features, which is not so discriminative at indistinct transition areas or fine edges. Consequently, their performance are sensitive to the size of unknown areas and may suffer from fuzzy boundaries and color blending. To address this issue, deep convolutional neural network (CNN)-based matting methods have been proposed\citep{xu2017deep,chen2018semantic,zhang2019late,Qiao_2020_CVPR,liu2020boosting,mgmatting} to leverage their strong representative abilities to learn discriminative features~\citep{zhang2020empowering}. Although CNN-based methods can achieve good matting results, the prerequisite trimaps or scribbles make them unlikely to be used in automatic applications such as the augmented reality of live streaming and film production. 

\begin{figure*}[!t]
\centering
\subfloat[]{\includegraphics[width=.425\linewidth]{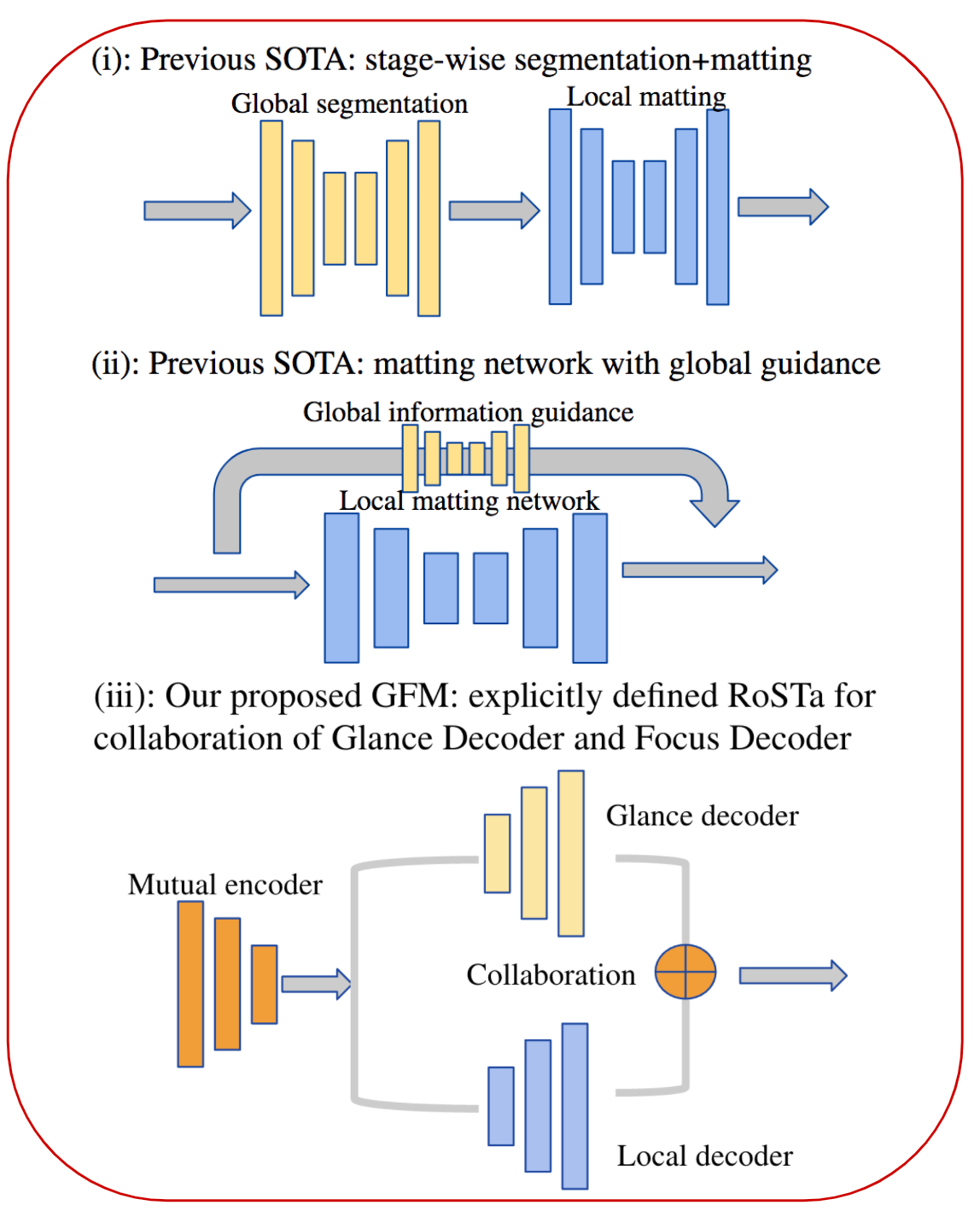}}
\hfil
\subfloat[]{\includegraphics[width=.575\linewidth]{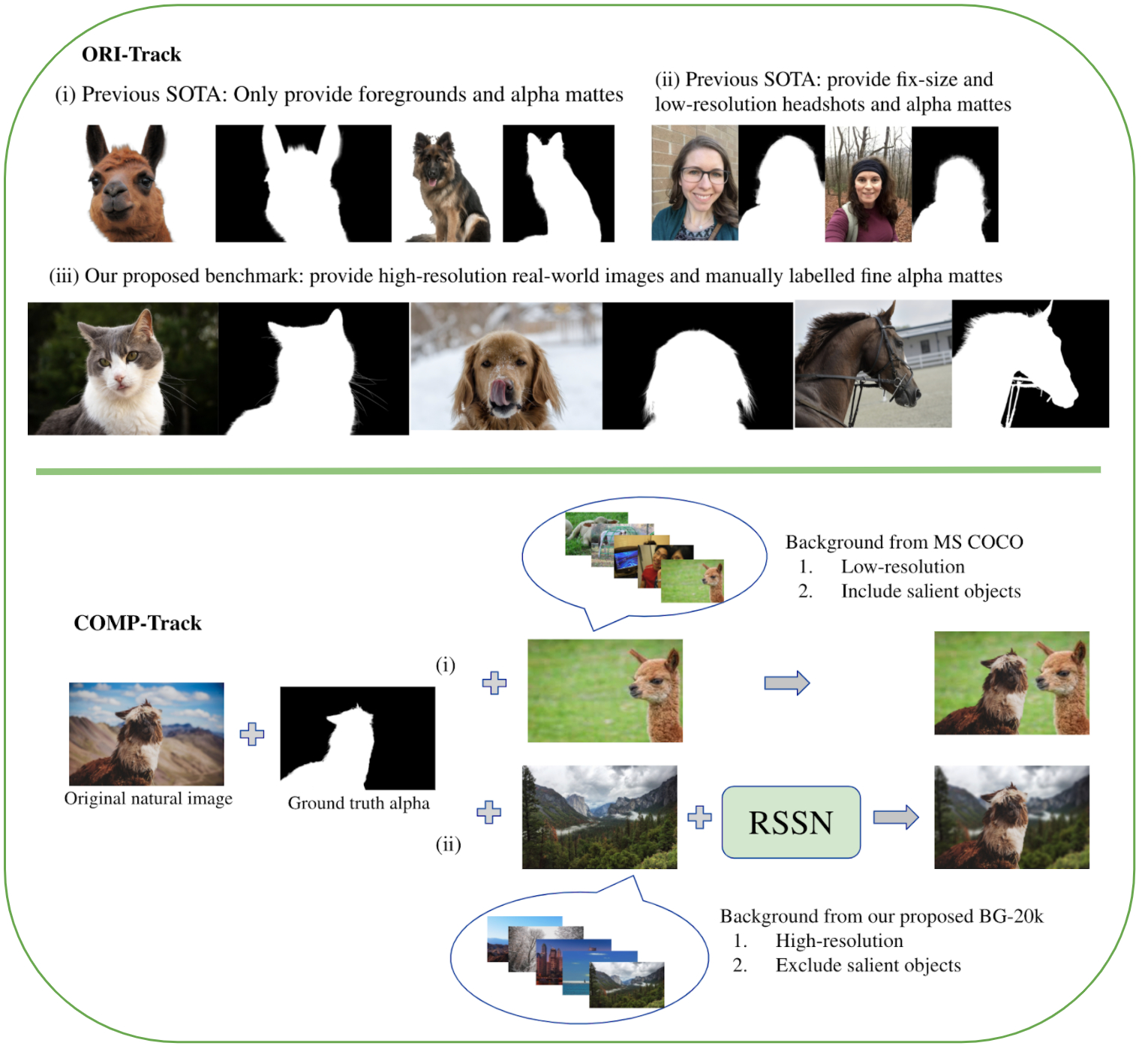}}
\caption{(a) The comparison between representative end-to-end-matting methods in (i) and (ii) and our GFM in (iii). (b) The comparison between existing matting datasets and our proposed benchmark as well as the comparison between existing composition methods and our RSSN.}
\label{fig:motivation}
\end{figure*}

To address this issue, end-to-end matting methods have been proposed~\citep{chen2018semantic,zhang2019late,shen2016deep,Qiao_2020_CVPR,liu2020boosting} in recent years. Most of them can be categorized into two types. The first type shown in (i) of Figure~\ref{fig:motivation}(a) is a straightforward solution, which is to perform global segmentation~\citep{aksoy2018semantic} and local matting sequentially. The former aims at trimap generation~\citep{chen2018semantic,shen2016deep} or foreground/background generation~\citep{zhang2019late} while the latter is image matting based on the trimap or other priors generated from the previous stage. The shortage of such a pipeline attributes to its sequential nature, since it may generate an erroneous semantic error which could not be corrected by the subsequent matting step. Besides, the separate training scheme in two stages may lead to a sub-optimal solution due to the mismatch between them. The second type is shown in (ii) of Figure~\ref{fig:motivation} (a), global information is provided as guidance while performing local matting. For example, coarse alpha matte is generated and used in the matting networks in \citep{liu2020boosting} and in \citep{Qiao_2020_CVPR}, spatial- and channel-wise attention are adopted to provide global appearance filtration to the matting network. Such methods avoid the problem of stage-wise modeling and training but bring in new problems. Although global guidance is provided in an implicit way, it is challenging to generate alpha matte for both foreground/background areas and transition areas simultaneously in a single network due to their distinct appearance and semantics. 

To solve the above problems, we study the distinct roles of semantics and details for natural image matting and explore the idea of decomposing the task into two parallel sub-tasks, semantic segmentation, and details matting. Specifically, we propose a novel end-to-end matting model named Glance and Focus Matting network (GFM). It consists of a shared encoder and two separate decoders to learn both tasks in a collaborative manner for natural image matting, which is trained end-to-end in a single stage. Moreover, we also explore different data representation formats in the Glance Decoder and gain useful empirical insights in the semantic-transition representations. As shown in Figure~\ref{fig:motivation}(a)(iii), compared with previous methods, GFM is a unified model that models both sub-tasks explicitly and collaboratively in a single network.

Another challenge for image matting is the limitation of the available matting datasets. As shown in Figure~\ref{fig:motivation}(b) ORI-Track, due to the laborious and costly labeling process, existing public matting datasets only have tens or hundreds of high-quality annotations~\citep{rhemann2009perceptually,shen2016deep,xu2017deep,zhang2019late,Qiao_2020_CVPR}. They either only provide foregrounds and alpha mattes~\citep{xu2017deep,Qiao_2020_CVPR} as in (i) of Figure~\ref{fig:motivation}(b) ORI-Track, or provide fix-size and low-resolution ($800\times600$) portrait images with inaccurate alpha mattes~\citep{shen2016deep} generated by an ensemble of existing matting algorithms as in (ii) of Figure~\ref{fig:motivation}(b) ORI-Track. Due to the unavailability of real-world original images, as shown in (i) of Figure~\ref{fig:motivation}(b) COMP-Track, a common practice for data augmentation in matting is to composite one foreground with various background images by alpha blending according to Eq.~\eqref{equa:blending} to generate large-scale composite data. The background images are usually choosing from existing benchmarks for image classification and detection, such as MS COCO \citep{lin2014microsoft} and PASCAL VOC \citep{everingham2010pascal}. However, these background images are in low-resolution and may contain salient objects. In this paper, we point out that the training images following the above route have a significant domain gap with those natural images due to the \emph{composition artifacts}, attributing to the resolution, sharpness, noise, and illumination discrepancies between the foreground and background images. The artifacts serve as cheap features to distinguish the foregrounds from the backgrounds and will mislead the models during training, resulting in overfitted models with poor generalization on natural images.

In this paper, we investigate the domain gap systematically and carry out comprehensive empirical analyses of the composition pipeline in image matting. We identify several kinds of discrepancies that lead to the domain gap and point out possible solutions to them. We then design a novel composition route named RSSN that can significantly reduce the domain gap arisen from the discrepancies of resolution, sharpness, noise, etc. Along with this, as shown in (ii) of Figure~\ref{fig:motivation}(b) COMP-Track, we propose a large-scale high-resolution clean background dataset (BG-20k) without salient foreground objects, which can be used in generating high-resolution composite images. Extensive experiments show that the proposed composition route along with BG-20k can reduce the generalization error by 60\% and achieve comparable performance as the model trained on original natural images. It opens an avenue for composition-based image matting since obtaining foreground images and alpha mattes are much easier than those from original natural images by leveraging chroma keying.

To fairly evaluate matting models' generalization ability on real-world images, we make the first attempt to establish a large-scale benchmark consists of 2,000 high-resolution real-world animal images and 10,000 real-world portrait images along with manually carefully labeled fine alpha mattes. Comparing with previous datasets~\citep{xu2017deep,Qiao_2020_CVPR,shen2016deep} as in (i) and (ii) of Figure~\ref{fig:motivation}(b) ORI-Track which only provide foreground images or low-resolution inaccurate alpha mattes, our benchmark includes all the high-resolution real-world original images and high-quality alpha mattes (more than 1080 pixels in the shorter side), which is beneficial to train models with better generalization on real-world images, and also suggests several new research problems which will be discussed later.

The contributions of this paper are four-fold:
\footnote{The source code, datasets, models, and a video demo will be made publicly available at \url{https://github.com/JizhiziLi/GFM}.}

$\bullet$ We propose a novel model named GFM for end-to-end image matting, which simultaneously generates global semantic segmentation and local alpha matte without any priors as input but a single image.

$\bullet$ We design a novel composition route RSSN to reduce various kinds of discrepancies and propose a large-scale high-resolution background dataset BG-20k to serve as better candidates for generating high-quality composite images.

$\bullet$ We construct a large-scale real-world images benchmark to benefit training a better model with good generalization by its large scale, diverse categories, and high-quality annotations.

$\bullet$ Extensive experiments on the benchmark demonstrate that GFM outperforms state-of-the-art (SOTA) matting models and can be a strong baseline for future research. Moreover, the proposed composition route RSSN demonstrates its value by reducing the generalization error by a large margin.


\section{Related Work}

\subsection{Image Matting} 
Most classical image matting methods are using auxiliary inputs like trimaps~\citep{li2017patch,sun2004poisson,levin2008spectral,chen2013knn,levin2007closed}. They sample or propagate foreground and background labels to the unknown areas based on local smoothness assumptions. Recently, CNN-based methods improve them by learning discriminative features rather than relying on hand-crafted low-level color features~\citep{xu2017deep,lu2019indices,hou2019context,cai2019disentangled,tang2019learning}. Deep Matting~\citep{xu2017deep} employed an encoder-decoder structure to extract high-level contextual features. IndexNet~\citep{lu2019indices} focused on boundary recovery by learning the activation indices during down-sampling. However, trimap-based methods require user interaction, make them unlikely to be deployed in automatic applications. Recently, Chen et al.~\citep{chen2018semantic} proposed an end-to-end model that first predicted the trimap then carried out matting. Zhang et al.~\citep{zhang2019late} also devised a two-stage model that first segmented the foregrounds or the backgrounds and then refined them with a fusion net. Both methods separated the process of segmentation and matting into different stages, which may generate erroneous segmentation results that mislead the subsequent matting step. Qiao et al.~\citep{Qiao_2020_CVPR} employed spatial and channel-wise attention to integrate appearance cues and pyramidal features while predicting, however, the distinct appearance and semantics of foreground/background areas and transition areas brought a heavy burden to a single-stage network and limited the quality of alpha matte prediction. Liu et al.~\citep{liu2020boosting} proposed a network to perform human matting by predicting the coarse mask first, then adopting a refinement network to predict a more detailed one. Despite the necessity of stage-wise training and testing, a coarse mask was not enough for guiding the network to refine the details since the transition areas were not defined explicitly.

In contrast to previous methods, we devise a novel end-to-end matting model via multi-task learning, which addresses the segmentation and matting tasks simultaneously. It can learn both high-level semantic features and low-level structural features in a shared encoder, benefits the subsequent segmentation and matting decoders collaboratively. One close related work with ours is AdaMatting \citep{cai2019disentangled}, which also has a structure of a shared encoder and two decoders. There are several significant differences: 1) AdaMatting requires a coarse trimap as an extra input while our GFM model only takes a single image as input without any priors; 2) the trimap branch in AdaMatting aims to refine the input trimap, which is much easier than generating a global representation in our case because the initial trimap actually serves as an attention mask for learning semantical features; 3) the adapted trimap generated from AdaMatting is serving as a guidance for alpha decoder via the following propagation unit, which is not suitable for end-to-end matting task since it lacks explicit collaboration for both decoders; 4) both the encoder and decoder structures of GFM are specifically designed for end-to-end matting, which differs from AdaMatting; and 5) we systematically investigate the semantic-transition representations in the glance decoder and gain useful empirical insights.

\subsection{Matting Dataset} 
Existing matting datasets~\citep{rhemann2009perceptually,xu2017deep,zhang2019late,Qiao_2020_CVPR} only contain foregrounds and a small number of annotated alpha mattes, $e.g.$, 27 training images and 8 test images in alphamatting~\citep{rhemann2009perceptually}, 431 training images and 50 test images in Comp-1k~\citep{xu2017deep}, and 596 training images and 50 test images in HAttMatting~\citep{Qiao_2020_CVPR}. DAPM~\citep{shen2016deep} proposes 2,000 real-world portrait images but at fix-size and low-resolution, together with limited quality alpha mattes generated by an ensemble of existing matting models. In contrast to them, we propose a high-quality benchmark consists of 10,000 high-resolution real-world portrait images and 2,000 animal images and manually annotated alpha matte for each image. We empirically demonstrate that the model trained on our benchmark has a better generalization ability on real-world images than the one trained on composite images.

\begin{figure*}[htbp]
    \includegraphics[width = 1\linewidth]{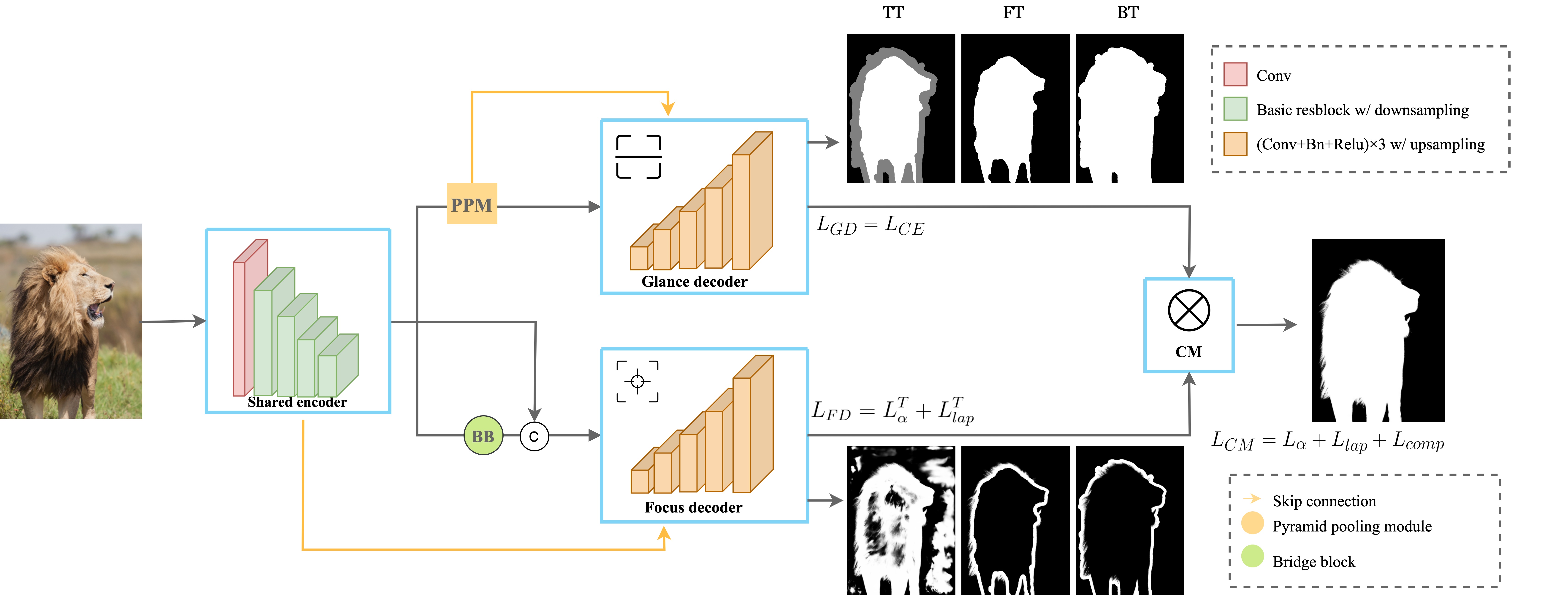}
    \caption{Diagram of the proposed Glance and Focus Matting (GFM) network, which consists of a shared encoder and two separate decoders responsible for rough segmentation of the whole image and details matting in the transition area.}
    \label{fig:network}
\end{figure*}

\subsection{Image Composition} 
As the inverse problem of image matting and the typical way of generating synthetic dataset, image composition plays an important role in image editing. Researchers have been dedicated to improve the reality of composite images from the perspective of color, lighting, texture compatibility, and geometric consistency in the past years \citep{xue2012understanding,dih,towardcomposition,dihdataset}. Xue et al.~\citep{xue2012understanding} conducted experiments to evaluate how the image statistical measures including luminance, color temperature, saturation, local contrast, and hue determined the realism of a composite. Tsai et al.~\citep{dih} proposed an end-to-end deep convolutional neural network to adjust the appearance of the foregrounds and backgrounds to be more compatible. Chen et al.~\citep{towardcomposition} proposed a generative adversarial network(GAN) architecture to learn geometrically and color consistent in the composites. Cong et al.~\citep{dihdataset} contributed a large-scale image harmonization dataset and a network using a novel domain verification discriminator to reduce the inconsistency of the foregrounds and the backgrounds. Although they did a good job in harmonizing the composites to be more realistic, the domain gap still exists when fitting the synthetic data into the matting model, the reason is that a subjective agreed standard of harmonization by a human is not equivalent to a good training candidate for a machine learning model. Besides, such procedures may modify the boundary of the foregrounds and result in inaccuracy of the ground truth alpha mattes. In this paper, we alternatively focus on generating composite images that can be used to reduce the generalization error of matting models on natural images.


\section{GFM: Glance and Focus Matting Network}

When tackling the image matting problem, we humans first glance at the image to quickly recognize the salient rough foreground or background areas and then focus on the transition areas to distinguish details from the background. It can be formulated as a segmentation stage and a matting stage roughly. Note that these two stages may be intertwined that there will be feedback from the second stage to correct the erroneous decision at the first stage, like some ambiguous areas caused by the protective coloration of animals or occlusions. To mimic the human experience and empower the matting model with proper abilities at both stages, it is reasonable to integrate them into a single model and explicitly model the collaboration. To this end, we propose a novel Glance and Focus Matting network for end-to-end image matting as shown in Figure~\ref{fig:network}.

\subsection{Shared Encoder}
GFM has an encoder-decoder structure, where the encoder is shared by two subsequent decoders. As shown in Figure~\ref{fig:network}, the encoder takes a single image as input and processes it through five blocks $E_0\sim E_4$, where each reduces the resolution by half. We adopt the DenseNet-121~\citep{huang2017densely}, ResNet-34, or ResNet-101~\citep{he2016deep} pre-trained on the ImageNet training set as our backbone encoder. Specifically, for DenseNet-121, we add a convolution layer to reduce the output feature channels to 512.

\subsection{Glance Decoder (GD)}
The glance decoder aims to recognize the easy semantic parts and leave the others as unknown areas. To this end, the decoder should have a large receptive field to learn high-level semantics. As shown in Figure~\ref{fig:network}, we symmetrically stack five blocks $D^G_4 \sim D^G_0$ as the decoder, each of which consists of three sequential $3\times3$ convolutional layers and an upsampling layer. To enlarge the receptive field further, we add a \textit{pyramid pooling module (PPM)}~\citep{zhao2017pyramid,Liu2019PoolSal} after $E_4$ to extract global context, which is then connected to each decoder block $D^G_i$ via element-wise sum. 

\textbf{Loss Function} The training loss for the glance decoder is a cross-entropy loss $L_{CE}$ defined as follows:
\begin{equation}
L_{CE} = -\sum_{c=1}^{C}{G_g^c}log\left(G_p^c\right),
\label{equa:ce_loss}
\end{equation}
where $G_p^c \in \left[0,1\right]$ is the predicted probability for $c$th class, $G_g^c \in \left\{0,1\right\}$ is the ground truth label. The output of GD is a two- or three-channel ($C=2$ or $3$) class probability map depends on the semantic-transition representation, which will be detailed in Section~\ref{section:data_presentation}.

\subsection{Focus Decoder (FD)}
As shown in Figure~\ref{fig:network}, FD has the same basic structure as GD, $i.e.$, symmetrically stacked five blocks $D^F_4 \sim D^F_0$. Different from GD, which aims to do roughly semantic segmentation, FD aims to extract details in the transition areas where low-level structural features are very useful. Therefore, we use a \textit{bridge block (BB)} \citep{Qin_2019_CVPR} instead of the PPM after $E_4$ to leverage local context in different receptive fields. Specifically, it consists of three dilated convolutional layers. The features from both $E_4$ and BB are concatenated and fed into $D^F_4$. We follow the U-net \citep{ronneberger2015u} style and add a shortcut between each encoder block $E_i$ and the decoder block $D^F_i$ to preserve fine details.

\textbf{Loss Function} The training loss for FD ($L_{FD}$) is composed of an alpha-prediction loss $L_{\alpha}^T$ and a Laplacian loss $L_{lap}^T$ in the unknown transition areas \citep{hou2019context}, $i.e.$,
\begin{equation}
L_{FD} = L_{\alpha}^T + L_{lap}^T.
\end{equation}
Following~\citep{xu2017deep}, the alpha loss $L_{\alpha}^T$ is calculated as absolute difference between ground truth $\alpha$ and predicted alpha matte $\alpha^{F}$ in the unknown transition region. It is defined as follows:
\begin{equation}
L_{\alpha}^T = \frac{\sum_{i}\sqrt{\left ( \left ( \alpha_{i}-\alpha_{i}^{F} \right )\times W^{T}_{i} \right )^{2}+\varepsilon^{2}}}{\sum_{i}W^{T}_{i}},
\label{equa:alpha_t}
\end{equation}
where $i$ denotes pixel index, $W^{T}_{i}\in \left \{ 0,1 \right \}$ denotes whether pixel $i$ belongs to the transition region or not. We add $\varepsilon=10^{-6}$ for computational stability. Following \citep{hou2019context}, the Laplacian loss $L_{lap}^T$ is defined as the $L1$ distance between the Laplacian pyramid of ground truth and that of prediction.
\begin{equation}
L_{lap}^T = {\sum_{i}W^{T}_{i}}\sum_{k=1}^{5}\left \|(Lap^k({\alpha_i})-Lap^k({\alpha^{F}_i}) \right \|_1,
\label{equa:lap_t}
\end{equation}
where $Lap^k$ denotes the $k$th level of the Laplacian pyramid. We use five levels in the Laplacian pyramid.

\subsection{RoSTa: Representation of Semantic and Transition Areas}
\label{section:data_presentation}
To investigate the impact of the representation format of the supervisory signal in our GFM, we adopt three kinds of \textbf{R}epresentations \textbf{o}f \textbf{S}emantic and \textbf{T}ransition \textbf{a}reas (RoSTa) as the bridge to link GD and FD.

\begin{itemize}
	\item \textbf{GFM-TT} We use the classical 3-class trimap $T$ as the supervisory signal for GD, which is generated by dilation and erosion from ground truth alpha matte with a kernel size of 25. We use the ground truth alpha matte $\alpha$ in the unknown transition areas as the supervisory signal for FD.
	\item \textbf{GFM-FT} We use the 2-class foreground segmentation mask $F$ as the supervisory signal for GD, which is generated by the erosion of ground truth alpha matte with a kernel size of 50 to ensure the left foreground part is correctly labeled. In this case, the area of $\mathcal{I}\left( \alpha > 0 \right)-F$ is treated as the transition area, where $\mathcal{I}\left( \cdot \right)$ denotes the indicator function. We use the ground truth alpha matte $\alpha$ in the transition area as the supervisory signal for FD.
	\item \textbf{GFM-BT} We use the 2-class background segmentation mask $B$ as the supervisory signal for glance decoder, which is generated by dilation of ground truth alpha matte with kernel size as 50 to ensure the left background part is correctly labeled. In this case, the area of $B - \mathcal{I}\left( \alpha > 0 \right)$ is treated as the transition area. We use the ground truth alpha matte $\alpha$ in the transition area as the supervisory signal for FD.
\end{itemize}

\subsection{Collaborative Matting (CM)}
As shown in Figure~\ref{fig:network}, CM merges the predictions from GD and FD to generate the final alpha prediction. Specifically, CM follows different rules when using different RoSTa as described in Section~\ref{section:data_presentation}. In GFM-TT, CM replaces the transition area of the prediction of GD with the prediction of FD. In GFM-FT, CM adds the predictions from GD and FD to generate the final alpha matte. In GFM-BT, CM subtracts the prediction of FD from the prediction of GD as the final alpha matte. In this way, GD takes charge of recognizing rough foreground and background by learning global semantic features, and FD is responsible for matting details in the unknown areas by learning local structural features. Such task decomposition and specifically designed parallel decoders make the model simpler than the two-stage ones in \citep{chen2018semantic,zhang2019late}. Besides, both decoders are trained simultaneously that the losses can be back-propagated to each of them via the CM module. In this way, our model enables interaction between both decoders, so that the erroneous prediction can be corrected in time by the responsible branch. Obviously, it is expected to be more effective than the two-stage framework, where the erroneous segmentation in the first stage could not be corrected by the subsequent one and thus mislead it.

\textbf{Loss Function} The training loss for collaborative matting ($L_{CM}$) consists of an alpha-prediction loss $L_{\alpha}$, a Laplacian loss $L_{lap}$, and a composition loss $L_{comp}$, $i.e.$,
\begin{equation}
L_{CM} = L_{\alpha} + L_{lap} + L_{comp}.
\end{equation}
Here $L_{\alpha}$ and $L_{lap}$ are calculated according to Eq.~\eqref{equa:alpha_t} and Eq.~\eqref{equa:lap_t} but in the whole alpha matte. Following \citep{xu2017deep}, the composition loss ($L_{comp}$) is calculated as the absolute difference between the composite images based on the ground truth alpha and the predicted alpha matte by referring to \citep{levin2007closed}. It can be defined as follows:
\begin{equation}
L_{comp} = \frac{\sum_{i}\sqrt{\left (C(\alpha_{i})-C(\alpha^{CM}_{i}) \right )^{2}+\varepsilon^{2}}} {N},
\end{equation}
where $C\left( \cdot \right)$ denotes the composited image, $\alpha^{CM}$ is the predicted alpha matte by CM, and $N$ denotes the number of pixels in the alpha matte.

To sum up, the final loss used during training is calculated as the sum of $L_{CE}$, $L_{FD}$ and $L_{CM}$, $i.e.$,
\begin{equation}
L = L_{CE} + L_{FD} + L_{CM}.
\end{equation}

\begin{figure*}[!t]
    \centering
    \includegraphics[width = .98\linewidth]{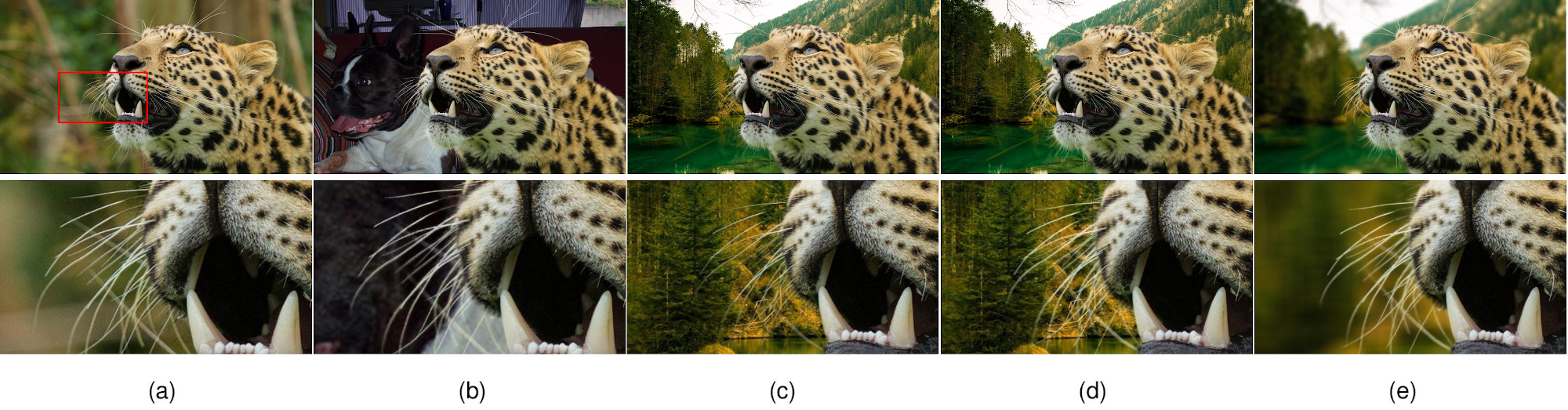}
     \caption{Comparison of different image composition methods. (a) Original natural image. (b) Composite with a background from MS COCO~\citep{lin2014microsoft} with the foreground computed by~\citep{levin2007closed}. (c) Composite with a background from our proposed BG-20k by alpha blending of original image directly. (d) Composite with background from our proposed BG-20k with the foreground computed by~\citep{levin2007closed}. (e) Composite with the large-aperture effect.}
    \label{fig:composite}
\end{figure*}

\section{RSSN: A Novel Composition Route}
 
Since labeling alpha matte of real-world natural images is very laborious and costly, a common practice is to generate large-scale composite images from a few foreground images and the paired alpha mattes \citep{xu2017deep}. The prevalent matting composition route is to paste one foreground with various background images by alpha blending according to Eq.~\eqref{equa:blending}. However, since the foreground and background images are usually sampled from different distributions, there will be a lot of \emph{composition artifacts} in the composite images, which leads to a large domain gap between the composite images and natural ones. The composition artifacts may mislead the model by serving as cheap features, resulting in overfitting on the composite images and producing large generalizing errors on natural images. In this section, we systematically analyze the factors that cause the \emph{composition artifacts} including \textbf{R}esolution discrepancy, \textbf{S}emantic ambiguity, \textbf{S}harpness discrepancy, and \textbf{N}oise discrepancy. To address these issues, we propose a new composition route named RSSN and a large-scale high-resolution background dataset named BG-20k.

\subsection{Resolution Discrepancy and Semantic Ambiguity}
\label{section: bg20k}
In the literature of image matting, the background images used for composition are usually chosen from existing benchmarks for image classification and detection, such as MS COCO \citep{lin2014microsoft} and PASCAL VOC \citep{everingham2010pascal}. However, these background images are in low-resolution and may contain salient objects, causing the following two types of discrepancies.

\begin{enumerate}
  \item \textbf{Resolution Discrepancy}: A typical image in MS COCO~\citep{lin2014microsoft} or Pascal VOC~\citep{everingham2010pascal} has a resolution about $389\times466$, which is much smaller compared to the high resolution foreground images in matting dataset such as Comp-1k~\citep{xu2017deep}. The resolution discrepancy between the foreground and background images will result in obvious artifacts as shown in Figure~\ref{fig:composite}(b).
  \item \textbf{Semantic ambiguity}: Images in MS COCO~\citep{lin2014microsoft} and Pascal VOC~\citep{everingham2010pascal} are collected for classification and object detection tasks, which usually contain salient objects from different categories, including various animals, human, and objects. Directly pasting the foreground image with such background images will result in semantic ambiguity for end-to-end image matting. For example, as shown in Figure~\ref{fig:composite}(b), there is a dog in the background which is beside the leopard in the composite image. Training with such images will mislead the model to ignore the background animal, $i.e.$, probably learning few about semantics but more about discrepancies.
\end{enumerate}

To address these issues, we collect a large-scale high-resolution dataset named BG-20k to serve as good background candidates for composition. We only selected those images whose shortest side has at least 1080 pixels to reduce the resolution discrepancy. Moreover, we removed those images containing salient objects to eliminate semantic ambiguity. The details of constructing BG-20k are presented as follows.

\begin{enumerate}
  \item We collected 50k high-resolution (HD) images using the keywords such as \textit{HD background, HD view, HD scene, HD wallpaper, abstract painting, interior design, art, landscape, nature, street, city, mountain, sea, urban, suburb} from websites with open licenses\footnote{\url{https://unsplash.com/} and \url{https://www.pexels.com/}}, removed those images whose shorter side has less than 1080 pixels and resized the left images to have 1080 pixels at the shorter side while keeping the original aspect ratio. The average resolution of images in BG-20k is $1180\times1539$;
  \item We removed duplicate images by a deep matching model~\citep{krizhevsky2012imagenet}. We adopted YOLO-v3~\citep{redmon2018yolov3} and an object detection method~\citep{chen2021recursive} to detect salient objects and then manually double-checked to make sure each image has no salient objects. In this way, we built BG-20k containing 20,000 high-resolution clean images;
  \item We split BG-20k into a disjoint training set (15k) and validation set (5k).
\end{enumerate}

\begin{figure}[htbp]
    \centering
    \includegraphics[width = .98\linewidth]{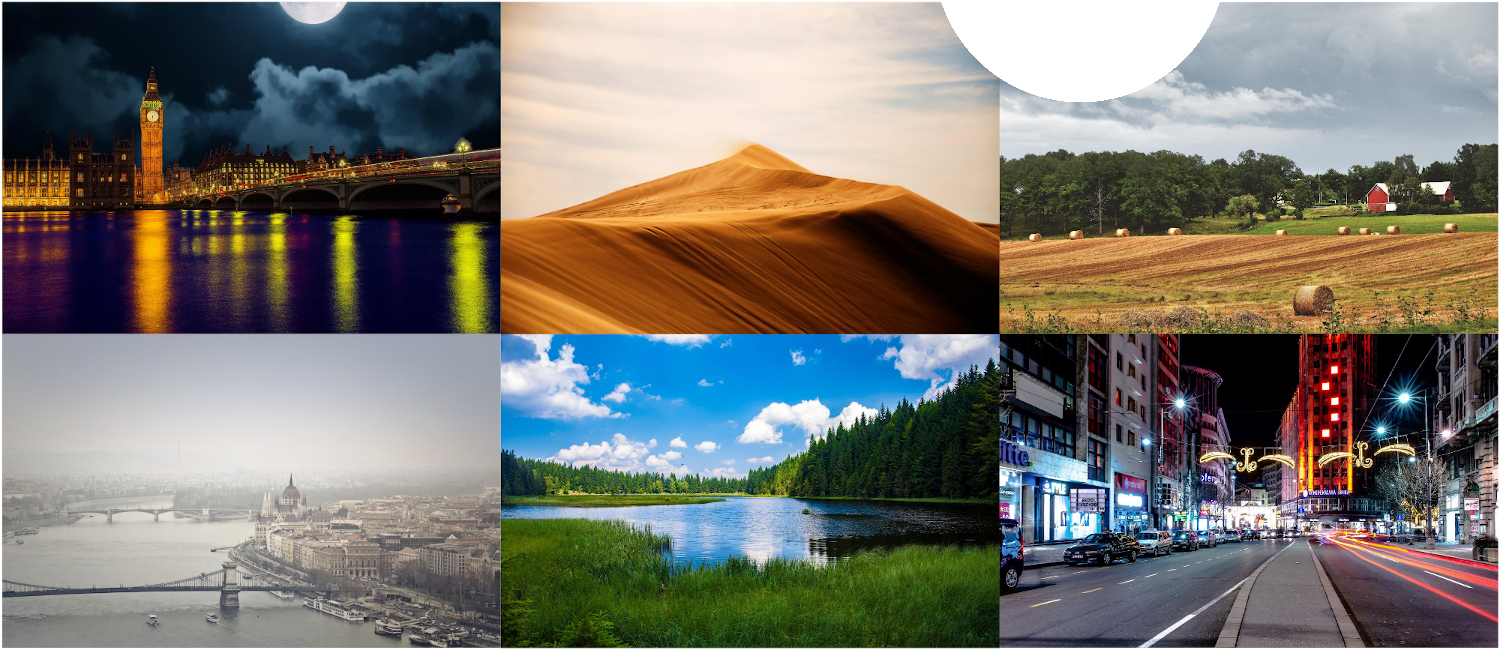}
     \caption{Some examples from our BG-20k dataset.}
    \label{fig:bg20k_examples}
\end{figure}

An composition example using the background image from BG-20k is shown in Figure~\ref{fig:composite}(c) and Figure~\ref{fig:composite}(d). In (c), we use the foreground image computed by multiplying the ground truth alpha matte with the original image for alpha blending, in (d), we use the foreground image computed by the method in \citep{levin2007closed} for alpha blending. As can be seen, there are obvious color artifacts in (c) that blend both colors of foreground and background in the fine details. The composite image in (d) is much more realistic than that in (c). Therefore, we adopt the method in \citep{levin2007closed} for computing the foreground images in our composition route. More examples of BG-20k are presented in Figure~\ref{fig:bg20k_examples} and the supplementary video.

\subsection{Sharpness Discrepancy}

In photography, it is very common to highlight the sharp and salient foreground from the background context by adopting a large aperture and focal length on the foreground with the shallow depth-of-field and blurring the background with the out-of-focus effect. An example is shown in Figure~\ref{fig:composite}(a), where the leopard is the center of interest and the background is blurred. Previous composition methods dismiss this effect, producing a domain gap of sharpness discrepancy between the composite images and natural photos. Since we target the image matting task, where the foregrounds are usually salient in the images, we thereby investigate this effect in our composition route. Specifically, we simulate it by adopting the averaging filter in OpenCV with a kernel size chosen from ${20,30,40,50,60}$ randomly to blur the background images. Since some natural photos may not have blurred backgrounds, we only use this technique in our composition route with a probability of 0.5. An example is shown in Figure~\ref{fig:composite}(e), where the background is chosen from BG-20k and blurred using the averaging filter. As can be seen, it has a similar style to the original image in (a).

\subsection{Noise Discrepancy}

Since the foregrounds and backgrounds come from different image sources, they may contain different noise distributions. This is another type of discrepancy, which will mislead the model to search noise cues during training, resulting in overfitting. To address this discrepancy, we adopt BM3D~\citep{dabov2009bm3d} to remove noise in both foreground and background images in RSSN. Furthermore, we add Gaussian noise with a standard deviation of 10 to the composite image such that the noise distributions in both foreground and background areas are the same. We find that it is effective in improving the generalization ability of trained models.

\subsection{The RSSN Composition Route}
\label{subsec:compositionroute}
We summarize the proposed composition route RSSN in Pipeline~\ref{alg:pipeline}. The input of the pipeline is the matting dataset, e.g., AM-2k and PM-10k as will be introduced in Section~\ref{sec:dataset}, DIM~\citep{xu2017deep}, or DAPM~\citep{shen2016deep}, and the proposed background image set BG-20k. If the matting dataset provides original images, e.g., AM-2K and PM-10k, we compute the foreground from the original image given the alpha matte by referring to~\citep{levin2007closed}. We random sample $K$ background candidates from BG-20k for each foreground for data augmentation. We set $K=5$ in our experiments. For each foreground image and background image, we carried out the denoising step with a probability of 0.5. To simulate the effect of large-aperture, we carried out the blur step on the background image with a probability of 0.5, where the blur kernel size was randomly sampled from $\{20,30,40,50,60\}$. We then generated the composite image according to the alpha-blending equation Eq.~\eqref{equa:blending}. Finally, with a probability of 0.5, we added Gaussian noise to the composite image to ensure the foreground and background areas have the same noise distribution. To this end, we generate a composite image set that has reduced many kinds of discrepancies, thereby narrowing the domain gap with natural images.

\begin{algorithm}
    \SetAlgorithmName{Pipeline}{}{}
\caption{The Proposed Composition Route: RSSN} 
\label{alg:pipeline}
\textbf{Input}: The matting dataset M containing $|M|$ images and
the background image set BG-20k  \\
\textbf{Output}: The composite image set C
\begin{algorithmic}[1]

\FOR{each $i \in [1,|M|]$}
\IF{there are original images in $M$, e.g. AM-2k, PM-10k }
\STATE {Sample an original image $I_i \in M$   } 
\STATE {Sample the paired alpha matte $\alpha_i \in M$   }
\STATE {Compute the foreground $F_i$ given $(I_i, \alpha_i)$ \citep{levin2007closed}}
\ELSE 
\STATE {Sample a foreground image $F_i \in M$   } 
\STATE {Sample the paired alpha matte $\alpha_i \in M$  } 
\ENDIF 

\FOR{each $k \in [1,K]$}
\STATE {Sample a background candidate $B_{ik} \in $ BG-20k  } 

\IF{$random()<0.5$}
\STATE { $F_i = Denoise(F_i)$ } $//$denoising by BM3D \citep{dabov2009bm3d}
\STATE { $B_{ik} = Denoise(B_{ik})$ }
\ENDIF 

\IF{$random()<0.5$}
\STATE {Sample a blur kernel size $r \in \{20,30,40,50,60\}$   }
\STATE { $B_{ik} = Blur(B_{ik}, r)$} $//$ the averaging filter
\ENDIF 

\STATE { Alpha blending: $C_{ik} = F_i \times \alpha_i + B_{ik}\times (1-\alpha_i)$}

\IF{$random()<0.5$}
\STATE { $C_{ik} = AddGaussianNoise(C_{ik})$}

\ENDIF 
\ENDFOR
\ENDFOR
\end{algorithmic}
\end{algorithm}

\section{Empirical Studies}
\label{sec:experiments}

\subsection{Benchmark for Real-world Image Matting}
\label{sec:dataset}
Due to the tedious process for generating manually labeled high-quality alpha mattes, the amount of real-world matting datasets is very limited, most previous methods adopted composite datasets such as Comp-1k~\citep{xu2017deep}, HATT-646~\citep{Qiao_2020_CVPR} and LF~\citep{zhang2019late} for data augmentation. However, as discussed in Section.~\ref{subsec:compositionroute}, the \emph{composition artifacts} caused by such convention would result in a large domain gap when adapting to real-world images. To fill this gap, we propose two large-scale high-resolution real-world image matting datasets AM-2k and PM-10k, consists of 2,000 animal images and 10,000 portrait images respectively, along with the high-quality manually labeled alpha mattes to serve as the appropriate training and testing bed for real-world image matting. We also set up two evaluation tracks for different purposes. Details are presented as follows.

\subsubsection{AM-2k}
\begin{figure}[htb]
    \centering
    \includegraphics[width = .98\linewidth]{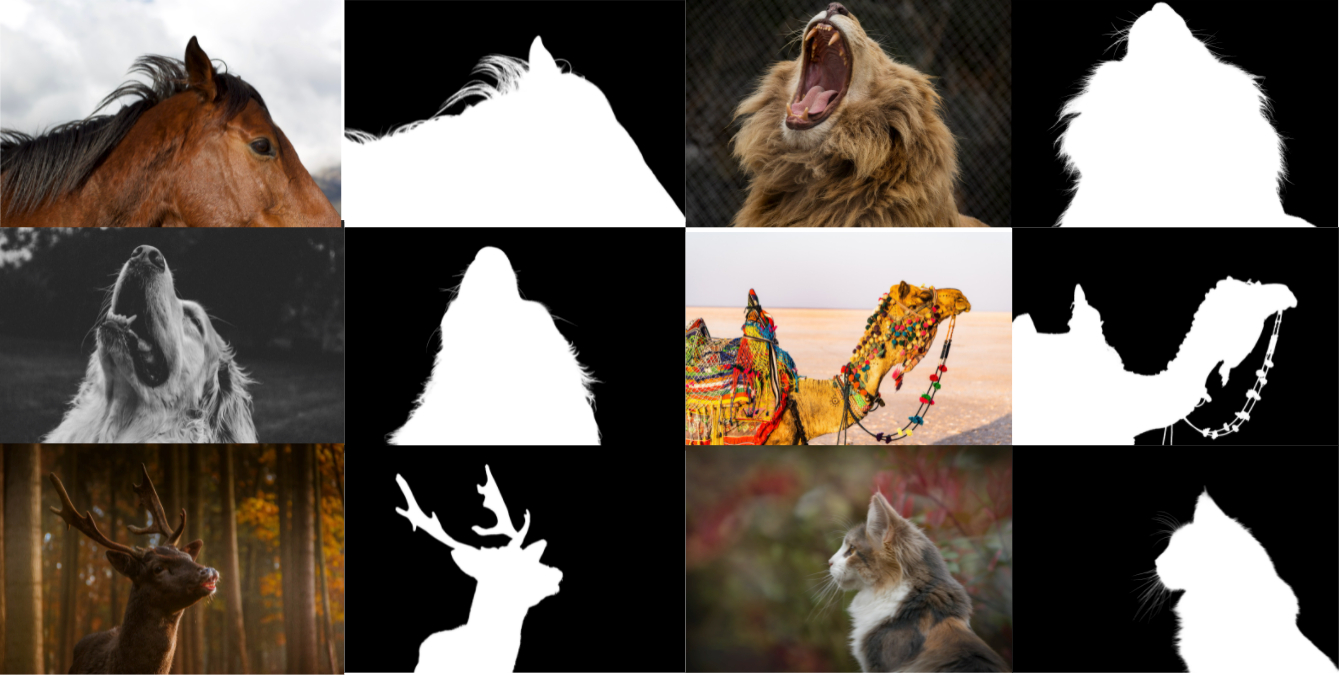}
     \caption{Some examples from our AM-2k dataset. The alpha matte is displayed on the right of the original image.}
    \label{fig:am2k_examples}
\end{figure}

AM-2k (Animal Matting 2,000 Dataset) consists of 2,000 high-resolution images collected and carefully selected from websites with open licenses. AM-2k contains 20 categories of animals including \textit{alpaca, antelope, bear, camel, cat, cattle, deer, dog, elephant, giraffe, horse, kangaroo, leopard, lion, monkey, rabbit, rhinoceros, sheep, tiger, zebra}, each with 100 real-world images of various appearance and diverse backgrounds. We ensure the shorter side of each image is more than 1080 pixels. We then manually annotate the alpha mattes using open-source image editing softwares, e.g., Adobe Photoshop, GIMP, etc. We randomly select 1,800 out of 2,000 to form the training set and the rest 200 as the validation set. Some examples and their ground truth are shown in Figure~\ref{fig:am2k_examples}. 

\subsubsection{PM-10k} PM-10k (Portrait Matting 10,000 Dataset) consists of 10,000 high-resolution images collected and carefully selected from websites with open licenses. We ensure PM-10k includes images with multiple postures and diverse backgrounds. We process the images by a human keypoint detection method~\citep{zhang2021towards} to ensure each image contains clear and salient human. We then generate the ground truth alpha mattes as in AM-2k. Finally, we split 9,500 of 10,000 to serve as the training set and 500 as the validation set. 

\subsubsection{Benchmark Tracks}
To benchmark the performance of matting models that 1) trained and tested both on real-world images; and 2) trained on composite images and tested on real-world images, we set up the following two evaluation tracks.

\paragraph{ORI-Track} (Original Images Based Track) is set to perform end-to-end matting tasks on the original real-world images. The ORI-Track is the primary benchmark track.

\paragraph{COMP-Track} (Composite Images Based Track) is set to investigate the influence of domain gap in image matting. As discussed before, the composite images have a large domain gap with natural images due to the composition artifacts. If we can reduce the domain gap and learn a domain-invariant feature representation, we can obtain a model with better generalization. To this end, we set up this track by making the first attempt towards this research direction. Specifically, we construct the composite training set by alpha-blending each foreground with five background images from the COCO dataset \citep{lin2014microsoft} (denoted as COMP-COCO) and our BG-20k dataset (denoted as COMP-BG20K), or adopting the composition route RSSN proposed in Section~\ref{subsec:compositionroute} based on our BG-20K (denoted as COMP-RSSN). Moreover, unlike previous benchmarks that evaluate matting methods on composite images~\citep{xu2017deep,zhang2019late,Qiao_2020_CVPR}, we evaluate matting methods on real-world images in the validation set same as the ORI-Track to validate their generalization ability.

Experiments were carried out on two tracks of the AM-2k and PM-10k datasets: 1) to compare the proposed GFM with SOTA methods, where we trained and evaluated them on the ORI-Track; and 2) to evaluate the side effect of domain gap caused by previous composition method and our proposed composition route, we trained and evaluated GFM and SOTA methods on the COMP-Track, $i.e.$, COMP-COCO, COMP-BG20K, and COMP-RSSN, respectively.

\begin{table*}[htbp]
\caption{Results on the ORI-Track and COMP-Track of AM-2k and PM-10k. $(d)$ stands for DenseNet-121~\citep{huang2017densely} backbone, $(r)$ stands for ResNet-34~\citep{he2016deep} backbone and $(r\dag)$ stands for ResNet-101~\citep{he2016deep} backbone. Representations of $TT$, $FT$ and $BT$ can refer to Section~\ref{section:data_presentation}.}
\resizebox{\linewidth}{!}{
\begin{tabular}{c|ccccc|ccc|ccc|ccc}
\hline
Dataset & \multicolumn{11}{|c}{\textbf{AM-2k}} \\
\hline
Track & \multicolumn{11}{|c}{ORI} \\
\hline
Method & SHM & LF & SSS &HATT &SHMC& GFM-TT(d) & GFM-FT(d) & GFM-BT(d)  & GFM-TT(r) & GFM-FT(r) & GFM-BT(r) &  GFM-TT(r$^\dag$) &GFM-FT(r$^\dag$) & GFM-BT(r$^\dag$) \\
\hline
SAD & 17.81 & 36.12 & 552.88 & 28.01 & 61.50 &10.26&12.74&12.74 &10.89&12.58&12.61 & \textbf{9.66}&11.54 & 12.80 \\
MSE & 0.0068 & 0.0116 & 0.2742 & 0.0055 & 0.0270 &0.0029&0.0038&0.0030 &0.0029&0.0037&0.0028 & \textbf{0.0024} &0.0032 & 0.0036\\
MAD & 0.0102 & 0.0210 & 0.3225 & 0.0161 & 0.0356 &0.0059&0.0075&0.0075 &0.0064&0.0073&0.0074 & \textbf{0.0056}& 0.0067& 0.0075 \\
Grad. & 12.54 & 21.06 & 60.81 & 18.29 & 37.00&\textbf{8.82} &9.98&9.13 &10.00&10.33&9.27 & 9.37 & 10.32 & 19.25\\
Conn. & 17.02 & 33.62 & 555.97 & 17.76 & 60.94 &9.57&11.78&10.07 &9.99&11.65&9.77 & \textbf{8.98} & 10.89 & 12.06\\
SAD-TRAN & 10.26 & 19.68 & 88.23 & 13.36 & 35.23 &\textbf{8.24}&9.66&8.67 &9.15&9.34&8.77 & 8.55 & 8.92 & 9.97\\
SAD-FG & 0.60 & 3.79 & 401.66 &1.36 & 10.93 &\textbf{0.42}&1.47&3.07 &0.77&1.31&2.84 & 0.53 & 0.92 & 0.93\\
SAD-BG & 6.95 & 12.55 & 62.99 &13.29 & 15.34 &1.59&1.61&1.00 &0.96&1.93&1.00 & \textbf{0.58} & 1.70 & 1.90\\
\hline
Track & \multicolumn{4}{|c}{COMP-COCO} & \multicolumn{4}{|c}{COMP-BG20K} & \multicolumn{6}{|c}{COMP-RSSN}\\
\hline
Method & SHM & \multicolumn{1}{c}{GFM-TT(d)} & \multicolumn{1}{c}{GFM-TT(r)} & \multicolumn{1}{c|}{GFM-TT(r$^\dag$)} &\multicolumn{1}{c}{SHM} &\multicolumn{1}{c}{GFM-TT(d)} & GFM-TT(r)& \multicolumn{1}{c|}{GFM-TT(r$^\dag$)} &SHM & GFM-TT(d)&\multicolumn{1}{c}{GFM-FT(d)} & GFM-BT(d)& GFM-TT(r) & GFM-TT(r$^\dag$)\\
\hline
SAD & 182.70&46.16&30.05&\multicolumn{1}{c|}{33.79} &\multicolumn{1}{c}{52.36} &\multicolumn{1}{c}{25.19}&16.44&\multicolumn{1}{c|}{15.88}&23.94&19.19 &\multicolumn{1}{c}{20.07}&22.82 &15.88&\textbf{14.78}\\
MSE &0.1017&0.0223&0.0129&\multicolumn{1}{c|}{0.0149}&\multicolumn{1}{c}{0.0268} &\multicolumn{1}{c}{0.0104}&0.0053&\multicolumn{1}{c|}{0.0048}&0.0099&\multicolumn{1}{c}{0.0069} &\multicolumn{1}{c}{0.0072}&0.0078 &0.0049&\textbf{0.0046}\\
MAD &0.1061&0.0273&0.0176&\multicolumn{1}{c|}{0.0192}&\multicolumn{1}{c}{0.0305} &\multicolumn{1}{c}{0.0146}&0.0096&\multicolumn{1}{c|}{0.0093}&0.0137&\multicolumn{1}{c}{0.0112}&\multicolumn{1}{c}{0.0118}&0.0133 &0.0092&\textbf{0.0086}\\
Grad. &64.74&20.75&17.22&\multicolumn{1}{c|}{16.63}&\multicolumn{1}{c}{22.87} &15.04&14.64&\multicolumn{1}{c|}{15.98}&17.66&\multicolumn{1}{c}{13.37} &\multicolumn{1}{c}{12.53}&12.49&14.04& \textbf{12.47}\\
Conn. &182.05&45.39&29.19&\multicolumn{1}{c|}{32.92}&\multicolumn{1}{c}{51.76} &24.31&15.57&\multicolumn{1}{c|}{15.17}&23.29&\multicolumn{1}{c}{18.31} &\multicolumn{1}{c}{19.08}&19.96&15.02&\textbf{14.10}\\
SAD-TRAN &25.01&17.10&15.21&\multicolumn{1}{c|}{14.09}&\multicolumn{1}{c}{15.32} &13.35&12.36&\multicolumn{1}{c|}{12.09}&12.63&\multicolumn{1}{c}{12.10} &\multicolumn{1}{c}{12.12}&12.06 &12.03&\textbf{11.20}\\
SAD-FG &23.26&8.71&4.74&\multicolumn{1}{c|}{5.36}&\multicolumn{1}{c}{3.52} &3.79&1.46&\multicolumn{1}{c|}{\textbf{1.38}}&4.56&\multicolumn{1}{c}{4.37} &\multicolumn{1}{c}{3.47}&5.20 &1.15& 2.11\\
SAD-BG &134.43&20.36&10.1&\multicolumn{1}{c|}{14.35}&\multicolumn{1}{c}{33.52} &8.05&2.62&\multicolumn{1}{c|}{2.41}&6.74&\multicolumn{1}{c}{2.72} &\multicolumn{1}{c}{4.48}&5.56 &2.71 & \textbf{1.47}\\
\hline 
Dataset & \multicolumn{11}{|c}{\textbf{PM-10k}} \\
\hline
Track & \multicolumn{11}{|c}{ORI} \\
\hline
Method & SHM & LF & SSS &HATT &SHMC& GFM-TT(d) & GFM-FT(d) & GFM-BT(d)  & GFM-TT(r) & GFM-FT(r) & GFM-BT(r) &GFM-TT(r$^\dag$) &GFM-FT(r$^\dag$) & GFM-BT(r$^\dag$) \\
\hline
SAD & 16.64 & 37.51 & 687.16& 22.66 &57.85  &11.89&12.76&13.45 &11.52&12.10& 13.34&\textbf{10.14}&12.36&12.76\\
MSE & 0.0069 & 0.0152& 0.3158 & 0.0038 & 0.0291 &0.0041&0.0044&0.0039&0.0038&0.0037& 0.0036&\textbf{0.0031}&0.0039 & 0.0039\\
MAD. & 0.0097 & 0.0152 &0.3958 & 0.0131 &  0.0340&0.0069&0.0074 & 0.0078&0.0067&0.0070 &0.0078&\textbf{0.0059}&0.0072 & 0.0074\\
Grad. & 14.54 & 21.82 & 69.72 & 15.16 & 37.28&12.90&\textbf{12.61}&13.22 &13.07&14.68& 13.09&12.88&21.19 & 25.13\\
Conn. & 16.13 & 36.92 & 691.08 & 11.95 &  57.86&11.24&12.15&11.37 &10.83&11.38& 10.54&\textbf{9.64}&11.89 & 12.17\\
SAD-TRAN & 8.53 & 16.36 & 63.64 & 9.32 &23.04 &7.80&7.81&7.80 &8.00&8.82& 8.02&\textbf{7.70}&8.37 & 8.83\\
SAD-FG & 0.74 & 11.63  & 240.02 &0.79 & 13.08 &1.65&\textbf{0.69}&3.98&0.97&0.93& 3.88&0.71&2.28 & 0.91\\
SAD-BG & 7.37 & 9.52 & 383.49 &12.54 & 21.72 &2.44&4.26&1.67&2.54&2.35& \textbf{1.44}&1.73&1.71 & 3.02\\
\hline
Track & \multicolumn{4}{|c}{COMP-COCO} & \multicolumn{4}{|c}{COMP-BG20K} & \multicolumn{6}{|c}{COMP-RSSN}\\
\hline
Method & SHM & \multicolumn{1}{c}{GFM-TT(d)} & \multicolumn{1}{c}{GFM-TT(r)} & \multicolumn{1}{c|}{GFM-TT(r$^\dag$)} &\multicolumn{1}{c}{SHM} &\multicolumn{1}{c}{GFM-TT(d)} & GFM-TT(r)& \multicolumn{1}{c|}{GFM-TT(r$^\dag$)} &SHM & GFM-TT(d)&\multicolumn{1}{c}{GFM-FT(d)} & GFM-BT(d)& GFM-TT(r) & GFM-TT(r$^\dag$)\\
\hline
SAD & 168.75&61.69&34.58 & \multicolumn{1}{c|}{33.90}& \multicolumn{1}{c}{34.06}&\multicolumn{1}{c}{21.54}&20.29&\multicolumn{1}{c|}{18.11}&22.02&\multicolumn{1}{c}{18.15} &\multicolumn{1}{c}{19.68}&21.80 &13.84 & \textbf{12.51}\\
MSE &0.0926 &0.0309&0.0165 &\multicolumn{1}{c|}{0.0157}& \multicolumn{1}{c}{0.0160}&\multicolumn{1}{c}{0.0088}&0.0086&\multicolumn{1}{c|}{0.0070}&0.0094&\multicolumn{1}{c}{0.0071} &\multicolumn{1}{c}{0.0078}& 0.0075&0.0049 & \textbf{0.0042}\\
MAD & 0.0960&0.0355&0.0198&\multicolumn{1}{c|}{0.0196} &\multicolumn{1}{c}{0.0194} &\multicolumn{1}{c}{0.0125}&0.0118&\multicolumn{1}{c|}{0.0106}&0.0126&\multicolumn{1}{c}{0.0106} &\multicolumn{1}{c}{0.0114}&0.0126 &0.0080 & \textbf{0.0072}\\
Grad. & 53.83&32.00&18.73 &\multicolumn{1}{c|}{25.26}& \multicolumn{1}{c}{23.02}&\multicolumn{1}{c}{19.21}&16.85&\multicolumn{1}{c|}{17.01}&18.65&\multicolumn{1}{c}{18.12} &\multicolumn{1}{c}{17.50}&16.97 &\textbf{14.44} & 14.62\\
Conn & 167.28&61.26&33.96 &\multicolumn{1}{c|}{33.54}&\multicolumn{1}{c}{33.7} &\multicolumn{1}{c}{20.97}&19.67&\multicolumn{1}{c|}{17.32}&21.61&\multicolumn{1}{c}{17.57} &\multicolumn{1}{c}{19.25}&19.09 &13.15 & \textbf{12.00}\\
SAD-TRAN &23.86 &17.69&11.21&\multicolumn{1}{c|}{12.95} &\multicolumn{1}{c}{12.85} &\multicolumn{1}{c}{11.82}&10.26&\multicolumn{1}{c|}{10.16}&10.6&\multicolumn{1}{c}{10.78} &\multicolumn{1}{c}{10.61}&10.38 &9.00 & \textbf{8.64}\\
SAD-FG &21.27 &17.42&13.99&\multicolumn{1}{c|}{8.18}&\multicolumn{1}{c}{9.66} &\multicolumn{1}{c}{5.31}&7.56&\multicolumn{1}{c|}{2.59}&5.24&\multicolumn{1}{c}{2.66} &\multicolumn{1}{c}{3.58}&5.74 &\textbf{1.32} & 1.83\\
SAD-BG & 123.62&26.59&9.37&\multicolumn{1}{c|}{12.77} &\multicolumn{1}{c}{11.55} &\multicolumn{1}{c}{4.41}&2.46&\multicolumn{1}{c|}{5.35}&6.19&\multicolumn{1}{c}{4.70} &\multicolumn{1}{c}{5.48}& 5.68&3.52 &\textbf{2.05}\\
\hline
\end{tabular}}
\label{tab:tracks_results}
\end{table*}

\subsection{Evaluation Metrics and Implementation Details}

\subsubsection{Evaluation Metrics} Following the common practice in \citep{rhemann2009perceptually,zhang2019late,xu2017deep}, we used the mean squared error (MSE), the sum of absolute differences (SAD), gradient (Grad.), and connectivity (Conn.) as the major metrics to evaluate the quality of alpha matte predictions. Note that the MSE and SAD metrics evaluate the pixel-wise differences between the prediction and the ground truth alpha matte, while the gradient and connectivity metrics favor clear details. Besides, we also use some auxiliary metrics such as Mean Absolute Difference (MAD), SAD-TRAN (SAD in the transition areas), SAD-FG (SAD in the foreground areas), and SAD-BG (SAD in the background areas) to comprehensively evaluate the quality of the alpha matte predictions. While MAD evaluates the average quantitative difference regardless of the image size, SAD-TRAN, SAD-FG, and SAD-BG evaluate SAD in different semantic areas, respectively. In addition, we compared the model complexity of different methods in terms of the number of parameters, computational complexity, and inference time.

\subsubsection{Implementation Details} During training, we used multi-scale augmentation similar to \citep{xu2017deep}. Specifically, we cropped each of the selected images with size from $\{640\times640, 960\times960, 1280\times1280\}$ randomly, resized the cropped image to $320\times320$, and randomly flipped it with a probability of 0.5. The encoder of GFM was initialized with DenseNet-121~\citep{huang2017densely}, ResNet-34, or ResNet-101~\citep{he2016deep} pre-trained on the ImageNet dataset. GFM was trained on two NVIDIA Tesla V100 GPUs. The batch size was 4 for DenseNet-121~\citep{huang2017densely}, 32 for ResNet-34, and 8 for ResNet-101~\citep{he2016deep}. For the COMP-Track, we composite five training images by using five different backgrounds for each foreground on-the-fly during training. It took about two days to train GFM for 500 epochs on ORI-Track and 100 epochs on COMP-Track. The learning rate was fixed to $1\times10^{-5}$ for both tracks.

For baseline end-to-end matting methods \textbf{LF}~\citep{zhang2019late} and \textbf{SSS}~\citep{aksoy2018semantic}, we used the official codes released by authors. For \textbf{SHM}~\citep{chen2018semantic}, \textbf{HATT}~\citep{Qiao_2020_CVPR} and \textbf{SHMC}~\citep{liu2020boosting} with no public codes, we re-implemented them according to the papers. For SHMC~\citep{liu2020boosting} which does not specify the backbone network, we used ResNet-34~\citep{he2016deep} for a fair comparison. These models were trained using the training set on the ORI-Track or COMP-Track.

Furthermore, we also evaluated the performance of several representative trimap-based matting methods, including \textbf{DIM}~\citep{xu2017deep}, \textbf{GCA}~\citep{li2020natural}, and \textbf{IndexNet}~\citep{hao2019indexnet} on both ORI-Track and COMP-Track of AM-2k and PM-10k. For DIM~\citep{xu2017deep}, we implement the code following the papers since the original code is not published. For GCA~\citep{li2020natural} and IndexNet~\citep{hao2019indexnet}, we used the official code released by the authors.

\subsection{Quantitative and Subjective Evaluation}

\subsubsection{Results on the ORI-Track} We benchmarked several SOTA methods~\citep{chen2018semantic,zhang2019late,aksoy2018semantic,Qiao_2020_CVPR,liu2020boosting} on the ORI-Track of AM-2k and PM-10k. The results are summarized in the top rows of Table~\ref{tab:tracks_results}. GFM-TT, GFM-FT, and GFM-BT denote the proposed GFM model with different RoSTa as described in Section~\ref{section:data_presentation}. $(d)$, $(r)$ and $(r\dag$) stand for using DenseNet-121~\citep{huang2017densely}, ResNet-34~\citep{he2016deep}, and ResNet-101~\citep{he2016deep} as the backbone encoder, respectively. There are several empirical findings from Table~\ref{tab:tracks_results}.

SSS~\citep{aksoy2018semantic} has a larger foreground and background SAD errors than other methods because it aims to extract all the semantic regions in the image rather than extracting the salient animal or portrait foregrounds like other methods. SHMC, which adopts global guidance~\citep{liu2020boosting} , and the stage-wise method LF~\citep{zhang2019late} perform better than SSS but have large SAD errors in the transition area. Because they have not explicitly defined the transition area, the matting networks have limited abilities to distinguish the details in the transition area. HATT~\citep{Qiao_2020_CVPR} obtains a smaller SAD error in the transition area and foreground area, owing to the attention module which can provide better global appearance filtration. SHM~\citep{chen2018semantic} performs better than the above methods, e.g., obtaining a smaller SAD error in the transition area and the background area than HATT~\citep{Qiao_2020_CVPR}. We believe the improvement credits to the explicit definition of RoSTa ($i.e.$, the trimap) and the PSPNet~\citep{zhao2017pyramid} used in the first stage which has a good semantic segmentation capability. Nevertheless, SHM still has a large error in the background area due to its stage-wise pipeline, which will accumulate the segmentation error into the matting network. 

Compared with all the SOTA methods, our GFM achieves the best performance by simultaneously segmenting the foregrounds, backgrounds, and matting on the transition areas, regardless of which RoSTa and encoder backbone it adopts. For example, it achieves the lowest SAD error in different areas, $i.e.$ 8.24 v.s. 10.26 for AM-2k, 7.70 v.s. 8.53 for PM-10k in the transition area, 0.42 v.s. 0.60 for AM-2k, 0.69 v.s. 0.74 for PM-10k in the foreground area, and 0.58 v.s. 6.95 for AM-2k, 1.44 v.s. 7.37 for PM-10k in the background area compared with the previous best method SHM~\citep{chen2018semantic}. The results of using different RoSTa are comparable, especially for FT and BT, since they both define two classes in the image for segmentation by the GD. GFM using TT as the RoSTa performs the best due to its explicit definition of the transition area as well as the foreground and background areas. We also tried three different backbone networks, DenseNet-121~\citep{huang2017densely}, ResNet-34, and ResNet-101~\citep{he2016deep}. All of them achieve the best performance compared with other SOTA methods. The superiority of GFM over other methods can be explained as follows. First, compared with stage-wise methods, GFM can be trained in a single stage and the collaboration module acts as an effective gateway to back-propagate matting errors to the responsible branch adaptively. Second, compared with those methods that adopt global guidance, GFM explicitly models the end-to-end matting task into two separate but collaborative sub-tasks by two distinct decoders. Moreover, it uses a collaboration module to merge the predictions according to the definition of RoSTa, which explicitly defines the role of each decoder.

\begin{figure*}[!t]
\centering
\includegraphics[width =.99\linewidth]{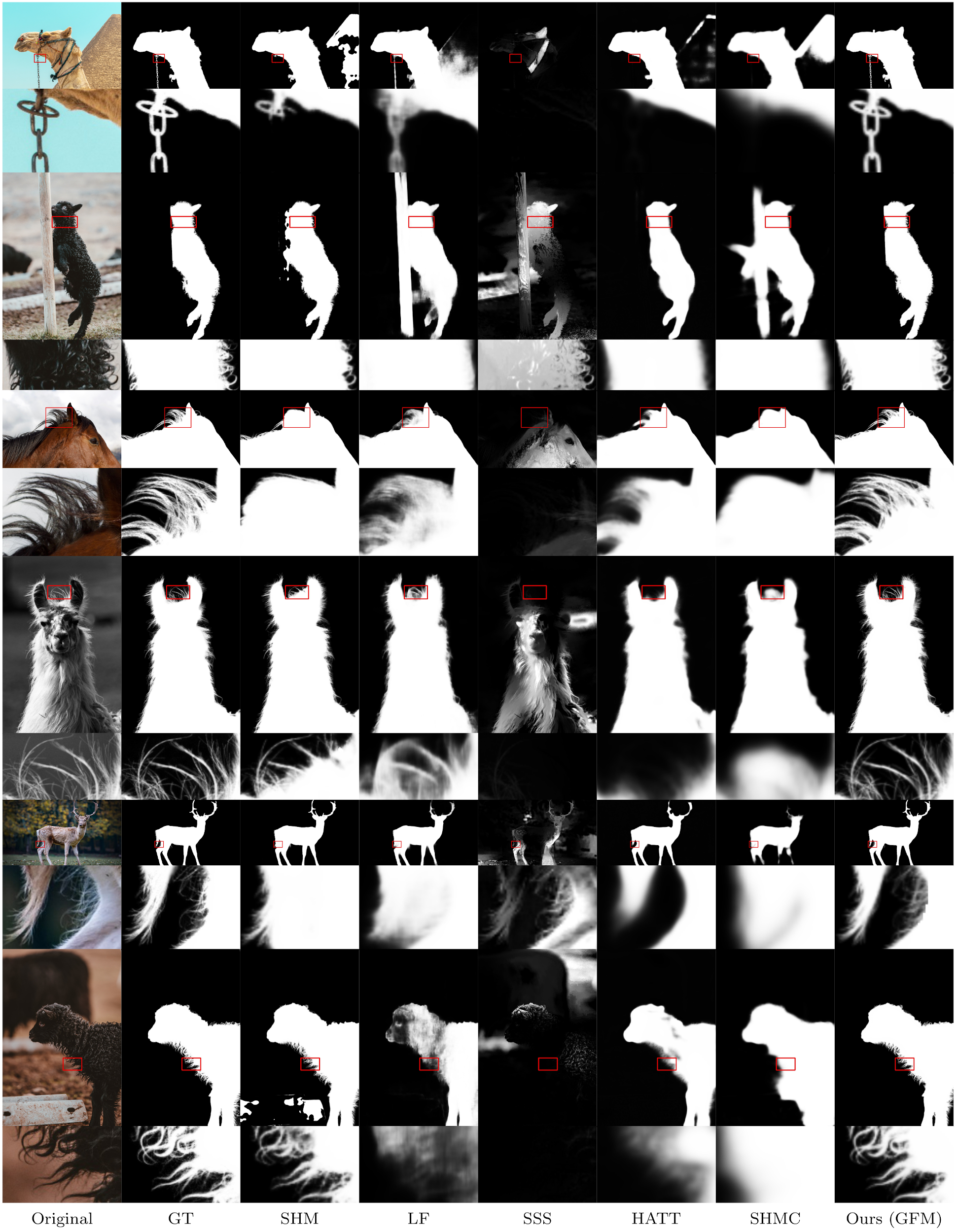}
    \caption{Subjective comparisons and the close views on AM-2k ORI-Track.}
    \label{fig:am2k_visual}
\end{figure*}
\begin{figure*}[!t]
\centering
\includegraphics[width =.99\linewidth]{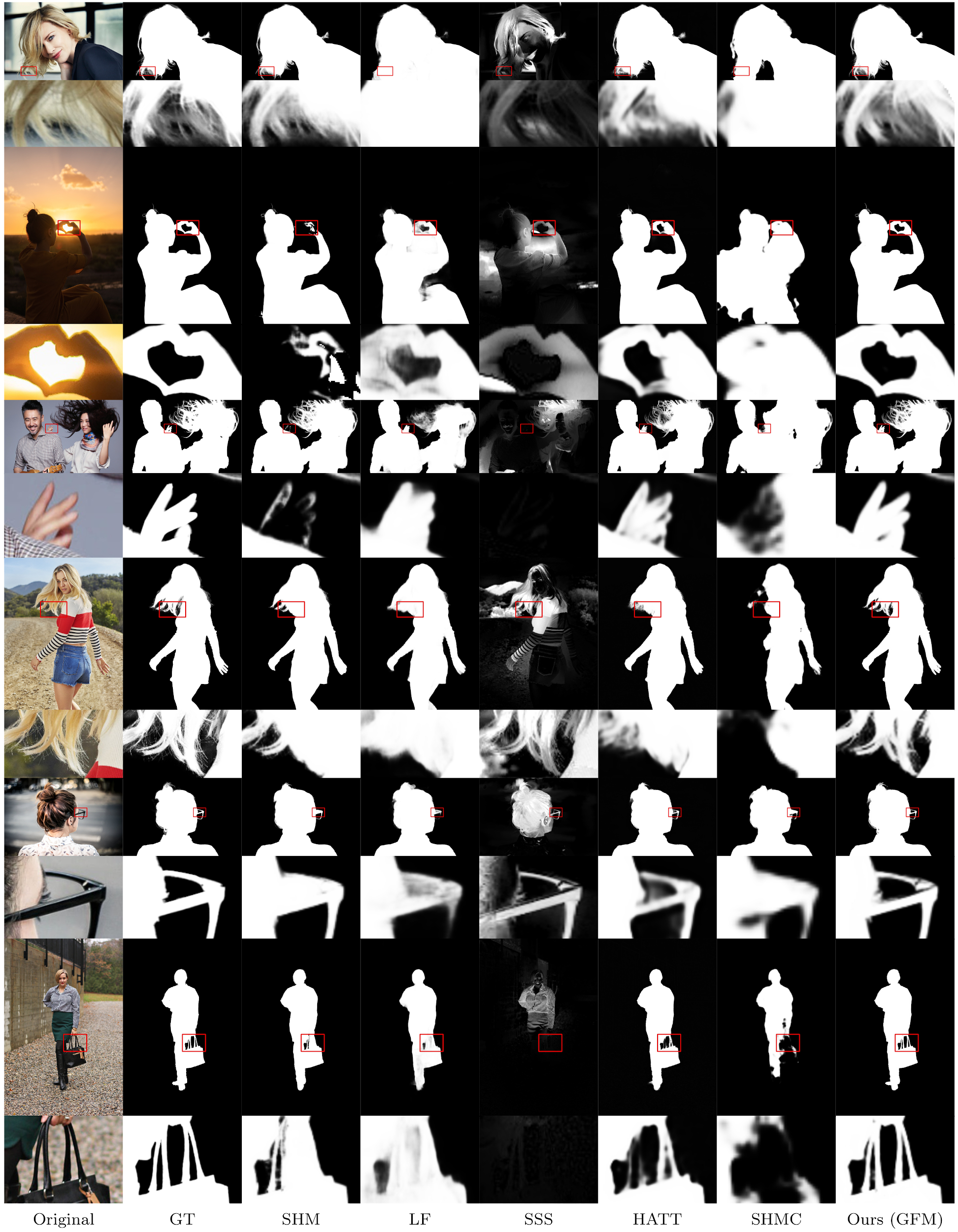}
    \caption{Subjective comparisons and the close views on PM-10k ORI-Track.}
    \label{fig:pm10k_visual}
\end{figure*}

\begin{table*}[htbp]
\caption{Results of trimap-based methods on the ORI-Track and COMP-Track of AM-2k and PM-10k.}
\resizebox{\linewidth}{!}{
\begin{tabular}{c|ccc|ccc|ccc|ccc}
\hline
Dataset & \multicolumn{11}{|c}{\textbf{AM-2k}} \\
\hline
Track & \multicolumn{3}{|c}{ORI} & \multicolumn{3}{|c}{COMP-COCO} & \multicolumn{3}{|c}{COMP-BG20K} & \multicolumn{3}{|c}{COMP-RSSN}\\
\hline
Method & DIM & GCA & IndexNet & DIM & GCA & IndexNet  & DIM & GCA & IndexNet & DIM & GCA & IndexNet \\
\hline
SAD & 6.82& 7.28& 7.40&7.59 & 8.97&9.59 & 7.45& 8.61&9.23 & 7.35&8.56 & 8.16 \\
MSE & 0.0010& 0.0011&0.0010 &0.0012 &0.0016 & 0.0015& 0.0011& 0.0015&0.0015 & 0.0011& 0.0015&0.0012 \\
MAD & 0.0040& 0.0042& 0.0043& 0.0044& 0.0052&0.0056 &0.0044 & 0.0050& 0.0054& 0.0043&0.0050  & 0.0047\\
Grad. &7.63 & 7.76& 8.81& 8.54&11.54& 12.93&8.48 & 10.92&12.71 & 8.46& 10.69& 9.86\\
Conn. & 6.17&6.50 & 6.41& 6.89& 8.40 &8.84 &6.75 & 8.02& 8.52& 6.68&7.92  &7.42\\
\hline 
Dataset & \multicolumn{11}{|c}{\textbf{PM-10k}} \\
\hline
Track & \multicolumn{3}{|c}{ORI} & \multicolumn{3}{|c}{COMP-COCO} & \multicolumn{3}{|c}{COMP-BG20K} & \multicolumn{3}{|c}{COMP-RSSN}\\
\hline
Method & DIM & GCA & IndexNet & DIM & GCA & IndexNet  & DIM & GCA & IndexNet & DIM & GCA & IndexNet \\
\hline
SAD & 6.00&6.16 & 6.44& 6.78& 7.45&7.36 &6.55 & 7.18&6.97 & 6.52& 6.30 &6.83\\
MSE & 0.0008&0.0008 & 0.0008&0.0010 &0.0012 &0.0011 &0.0010 & 0.0011&0.0010 &0.0010 & 0.0090 &0.0010\\
MAD & 0.0035&0.0036 &0.0038 & 0.0039& 0.0043&0.0043 &0.0038 & 0.0042& 0.0041& 0.0038&  0.0037&0.0040\\
Grad. &8.59 & 8.32&9.87 & 10.58& 10.45&10.98 & 9.96&10.07 & 10.06& 10.05& 7.43 &9.90\\
Conn. &5.55 & 6.16& 5.75& 6.32& 6.99&6.79 & 6.11& 6.71& 6.47& 6.08& 5.78 &6.24\\
\hline 
\end{tabular}}
\label{tab:trimap_based}
\end{table*}

From Figure~\ref{fig:am2k_visual} and Figure~\ref{fig:pm10k_visual}, we can find similar observations. SHM~\citep{chen2018semantic}, LF~\citep{zhang2019late}, and SSS~\citep{aksoy2018semantic} fail to segment some foreground parts, implying inferiority of its stage-wise network structure. HATT~\citep{Qiao_2020_CVPR} and SHMC~\citep{liu2020boosting} struggle to obtain clear details in the transition areas since the global guidance is helpful for recognizing the semantic areas while being less useful for matting of details. Compared to them, our GFM achieves the best results owing to the advantage of a unified model, which deals with the foreground/background and transition areas using separate decoders and optimizes them in a collaborative manner. More results of GFM can be found in the demo video.

\subsubsection{Results on the COMP-Track} 
\justifying{We evaluated SHM~\citep{chen2018semantic}, the best performed SOTA method, and our GFM with three different backbones on the COMP-Track of AM-2k and PM-10k including COMP-COCO, COMP-BG20K, and COMP-RSSN. The results are summarized in the bottom rows of Table~\ref{tab:tracks_results}, from which we have several empirical findings. \textbf{First}, when training matting models using images from MS COCO dataset~\citep{lin2014microsoft} as backgrounds, GFM performs much better than SHM~\citep{chen2018semantic}, $i.e$, 46.16, 30.05, and 33.79 v.s. 182.70 for AM-2k, 61.69, 34.58, and 33.90 v.s. 168.75 for PM-10k in terms of whole image SAD, confirming the superiority of the proposed model over the two-stage one for generalization. \textbf{Second}, GFM using ResNet-34 or ResNet-101~\citep{he2016deep} performs better than using DenseNet-121~\citep{huang2017densely}, implying that the residual structure in ResNet has better representation ability in extracting more accurate semantic representations and generalization ability in dealing with the domain gap between composite and real data, especially at the settings of COMP-BG20K Track and COMP-RSSN Track. \textbf{Third}, when training matting models using background images from the proposed BG-20k dataset, the errors of all the methods are significantly reduced, especially for SHM~\citep{chen2018semantic}, $i.e.$, from 182.70 to 52.36 for AM-2k, or 168.75 to 34.06 for PM-10k, which mainly attributes to the reduction of SAD error in the background area, $i.e.$, from 134.43 to 33.52 for AM-2k and 123.62 to 11.55 for PM-10k. There is the same trend for GFM(d), GFM(r) and GFM($r\dag$). These results confirm the value of our BG-20k, which helps to reduce resolution discrepancy and eliminate semantic ambiguity in the background area. }

\textbf{Fourth}, when using the proposed RSSN for training, the errors can be reduced further for SHM~\citep{chen2018semantic}, $i.e.$, from 52.36 to  23.94 for AM-2k and 34.06 to 22.02 for PM-10k, from 25.19 to 19.19 and 21.54 to 18.15 for GFM(d), from 16.44 to 15.88 and 20.29 to 13.84 for GFM(r), and from 15.88 to 14.78, 18.11 to 12.51 for GFM($r\dag$) The improvement is attributed to the composition techniques in RSSN: 1) we simulate the large-aperture effect to reduce sharpness discrepancy; and 2) we remove the noise of foregrounds/backgrounds and add noise to the composite images to reduce noise discrepancy. Note that the SAD error of SHM~\citep{chen2018semantic} has dramatically reduced about 87\% from 182.70 to 23.93 for AM-2k or 168.75 to 22.02 for PM-10k when using RSSN compared with the traditional composition method based on MS COCO dataset, which is even comparable with the one obtained by training using original images, $i.e.$, 17.81 for AM-2k and 16.64 for PM-10k. It demonstrates that the proposed composition route RSSN can significantly \textit{narrow} the domain gap and help to learn down-invariant features. \textbf{Last}, We also conducted experiments by using different RoSTa in GFM(d) on the track COMP-RSSN, their results have a similar trend to those on the ORI-TRACK.

\subsubsection{Results of Trimap-based Matting Methods} We also benchmarked several SOTA trimap-based matting methods~\citep{xu2017deep,li2020natural,hao2019indexnet} on the ORI-Track and COMP-Track of AM-2k and PM-10k. The results are summarized in the Table~\ref{tab:trimap_based}. As can be seen, the performance of trimap-based methods follows the same trends as the trimap-free matting methods on both tracks. For all the three methods, there are performance gaps between training them using the MS COCO \citep{lin2014microsoft} background and the original images from our dataset as background. The gaps are different for each method, $i.e.$ 7.59 to 6.82, 6.78 to 6.00 for DIM on the two datasets, 8.97 to 7.28, 7.45 to 6.16 for GCA, 9.59 to 7.40, 7.36 to 6.44 for IndexNet, respectively. As always, after employing BG-20k and RSSN, the error has decreased for all methods on two datasets. Besides, there are two points to note, 1) even with RSSN, the performance gaps still exist for trimap-based matting methods, which deserves further studies; and 2) trimap-based matting methods perform better than our GFM in terms of SAD-TRAN, $i.e.$, 6.82 to 8.24 on AM-2k, 6.00 to 7.70 on PM-10k, implying that GFM can be further improved on the transition area.

\subsubsection{GFM with Trimap-based Matting Method} To further investigate the effectiveness of GFM, we provide the evaluation results of several variants on the ORI-Track of AM-2k in Table~\ref{tab:gfm_with_tbase}, including 1) DIM: the trimap-based matting method DIM~\citep{xu2017deep}, which is trained and tested both on the ground truth trimap; 2) GFM(d): our proposed GFM with DesNet-121 as the backbone; 3) GFM(d)+DIM: we replace the predicted alpha matte in the transition area with the result from DIM, which is trained on the ground truth trimap while tested on the predicted trimap generated by GFM GD; and 4) SHM~\citep{chen2018semantic}: a two-stage "trimap prediction + trimap-based matting method" network, where its trimap-based matting method is very similar to DIM. As always, SAD, MSE, MAD, Grad., and Conn. are calculated on the whole image. SAD-TRAN, SAD-FG, and SAD-BG calculated in transition, foreground, and background area respectively.

\begin{table}[htbp] 
\caption{Comparison of GFM and its variant with DIM~\citep{xu2017deep} and SHM \citep{chen2018semantic} on the ORI-Track of AM-2k.}
\resizebox{\linewidth}{!}{
\begin{tabular}{l|c|cccc|c}
\hline
Method & DIM & GFM(d) & GFM(d)+DIM & SHM \\
\hline
SAD &  6.82 & 10.26 & 11.46 & 17.81 \\
MSE &  0.0010 & 0.0029 & 0.0032 & 0.0068 \\
MAD &  0.0040 & 0.0059 & 0.0066 & 0.0102 \\
Grad. &  7.63 & 8.82& 11.41 & 12.54 \\
Conn. & 6.17 & 9.57 &10.88 & 17.02 \\
SAD-TRAN & 6.82 & 8.24 & 9.54 &10.26 \\
SAD-FG &  0 & 0.42& 0.42 & 0.60 \\
SAD-BG & 0 & 1.59 &1.49 & 6.95 \\
\hline
\end{tabular}}
\label{tab:gfm_with_tbase}
\end{table}

There are several conclusions we can draw from Table~\ref{tab:gfm_with_tbase}. \textbf{First,} GFM still performs the best among all the trimap-free matting methods. Specifically, compared with the ``GFM+DIM'' variant, which uses the trimap-based matting method DIM~\citep{xu2017deep} to obtain the alpha matte in the transition area, our end-to-end GFM model still performs better, especially in the transition area, $i.e.$ 8.24 to 9.54 for GFM(d). These results validate the effectiveness of the FD in our GFM. \textbf{Second}, comparing DIM with ``GFM+DIM'' variant, we can find that DIM is very sensitive to the trimap, e.g., 6.82 to 9.54 for GFM(d). It can also explain the above performance gap between GFM and ``GFM+DIM'' variant. These results validate the effectiveness of our proposed one-stage ``sharing encoder + multi-task decoder'' structure since the FD can be adapted to the predicted trimap from the GD, owing to the collaboration matting module. This can be further proven by comparing GFM(d) with SHM~\citep{chen2018semantic}, which is a typical two-stage structure. \textbf{Third}, compared with trimap-based method DIM, which is trained and tested on the ground truth trimap, there are still rooms to improve GFM in the transition area (SAD 6.82 to 8.24), which can be the future work.

\subsection{Ablation Study}

\begin{table}[htb]
\caption{Ablation study of GFM on AM-2k.}
\begin{center}
\resizebox{\linewidth}{!}{
\begin{tabular}{l|ccccc}
\hline
Track & \multicolumn{5}{|c}{ORI} \\
\hline
Method & SAD & MSE & MAD & Grad & Conn \\
\hline
 GFM-TT(d) &10.26 & 0.0029 &\textbf{0.0059} &8.82 &9.57\\
 GFM-TT(r) &10.89 & 0.0029 &0.0064 &10.00 &9.99\\
GFM-TT(r2b) &\textbf{10.24} & \textbf{0.0028} &0.0060 &\textbf{8.65} &\textbf{9.33}\\
 GFM-TT-SINGLE(d) &13.79&0.0040 &0.0081 &13.45 &13.04\\
 GFM-TT-SINGLE(r) &15.50&0.0040 &0.0091 &14.21 &13.15\\
 GFM-TT(d) w/ one PPM & 10.71 & 0.0029 & 0.0063 & 10.23 & 9.88 \\
 GFM-TT(d) excl. PPM & 10.86 & 0.0030 &0.0064& 9.91 & 9.92 \\
 GFM-TT(d) excl. BB & 11.27 & 0.0035 &0.0067& 9.33 & 10.40 \\
 GFM-TT(r) excl. PPM & 11.90 & 0.0035 &0.0070& 10.50 & 11.07 \\
 GFM-TT(r) excl. BB & 11.29 & 0.0032 &0.0066& 9.59 & 10.43 \\

 \hline
Track & \multicolumn{5}{|c}{COMP-RSSN} \\
\hline
 GFM-TT(d) & 25.19 & 0.0104 &0.0146 &15.04 &24.31\\
 GFM-TT(d) w/ blur & 21.37 & 0.0081 &0.0124 &14.31&20.50\\
 GFM-TT(d) w/ denoise & 22.95 & 0.0090 &0.0134 &14.37 &22.10\\
 GFM-TT(d) w/ noise &19.87 & 0.0075 &0.0116&\textbf{13.22} &18.97\\
 GFM-TT(d) w/ RSSN & \textbf{19.19} & \textbf{0.0069} &\textbf{0.0112} &13.37 &\textbf{18.31}\\
 \hline
\end{tabular}}
\end{center}
\label{tab:gfm_ablation_study}
\end{table}

\subsubsection{Results on the ORI-Track} 
To further verify the benefit of the designed structure of GFM, we conducted ablation studies on several variants of GFM on the ORI-Track of AM-2k, including 1) motivated by Qin et.al ~\citep{Qin_2019_CVPR}, in GFM encoder when using ResNet-34~\citep{he2016deep} as the backbone, we modified the convolution kernel of $E_0$ from $7\times7$ with stride 2 to $3\times3$ with stride 1, removed the first max pooling layer in $E_0$, and added two more encoder layers $E_5$ and $E_6$ after $E_4$, each of which had a max pooling layer with stride 2 and three basic res-blocks with 512 filters, denoting ``r2b''; 2) using a single decoder to replace both FD and GD in GFM, denoting ``SINGLE''; 3) excluding the pyramid pooling module (PPM) in GD, and 4) excluding the bridge block (BB) in FD. The results are summarized in the top rows of Table~\ref{tab:gfm_ablation_study}. \textbf{First}, when using $r2b$ structure, all the metrics have been improved compared with GFM-TT(r), which is attributed to the larger feature maps at the early stage of the encoder part. However, it has more parameters and computations than GFM-TT(r), which will be discussed later. \textbf{Second}, using a single decoder results in worse performance, $i.e.$, SAD increases from 10.26 to 13.79 for GFM-TT(d) and 10.89 to 15.50 for GFM-TT(r), which confirms the value of decomposing the end-to-end image matting task into two collaborative sub-tasks. \textbf{Third}, without PPM, SAD increases from 10.26 to 10.86 for GFM-TT(d) and 10.89 to 11.90 for GFM-TT(r), demonstrating that the global context features by PPM due to its larger receptive field are beneficial for semantic segmentation in GD. Moreover, compared with excluding PPM, using one PPM block leads to better performance but still falls behind the default setting, $i.e.$, using five PPM blocks. \textbf{Fourth}, without BB, SAD increases from 10.26 to 11.27 for GFM-TT(d) and 10.89 to 11.29 for GFM-TT(r), demonstrating that the learned local structural features from BB are beneficial for matting in FD due to its dilated convolutional layers.

\subsubsection{Results on the COMP-Track} 
To verify the different techniques in the proposed composition route RSSN, we conducted ablation studies on several variants of RSSN, including 1) only using the simulation of large-aperture effect, denoting ``w/ blur''; 2) only removing foreground and background noise, denoting ``w/ denoise''; 3) only adding noise on the composite images, denoting ``w/ noise''; and 4) using all the techniques in RSSN, denoting ``w/ RSSN''. We used the BG-20k for sampling background images in these experiments. The results are summarized in the bottom rows of Table~\ref{tab:gfm_ablation_study}. \textbf{First}, compared with the baseline model listed in the first row, which was trained using the composite images by alpha-blending, each technique in the proposed composition route is helpful to improve the matting performance in terms of all the metrics. \textbf{Second}, simulation of large-aperture effect and adding noise on the composite images are more effective than denoising. \textbf{Third}, different techniques are complementary to each other in that they contribute to the best performance achieved by RSSN collaboratively.

\begin{table}[htbp]
\caption{Results of GFM using different RoSTa and two RoSTa integration techniques on AM-2k.}
\begin{center}
\resizebox{\linewidth}{!}{
\begin{tabular}{c|ccccc|c}
\hline
RoSTa  & SAD & MSE & MAD & Grad. & Conn. & Speed (s) \\
\hline
TT & 11.28 & 0.0030 & 0.0065&\textbf{10.89}&10.63 & 0.1536\\
BT & 11.76 & 0.0033&0.0070 &11.56&11.15 & 0.1416\\
FT & 12.26 &0.0034 & 0.0072&12.52&11.68 & 0.1326 \\
\hline
EN-median & \textbf{10.62}&\underline{0.0026} & \textbf{0.0062}&\underline{11.19}&\textbf{9.85} & 0.4644\\
GFM-RIM  & \underline{10.79} & \textbf{0.0025}&\underline{0.0063}&11.61& \underline{10.11}  & 0.1859\\
\hline
\end{tabular}}
\end{center}
\label{tab:ensemble}
\end{table}


\subsection{RoSTa Integration and Hybrid-resolution Test}

\subsubsection{RoSTa Integration}

Since TT, FT, and BT have their own advantages, how to design the network to make benefit from them remains a challenge. In this paper, we explored two RoSTa integration techniques on AM-2k, and compared them with individual RoSTa. Please note that for a fair comparison, we set the testing resolution ratio as $1/3$ for all the methods. The results are shown in Table~\ref{tab:ensemble}.

A straightforward solution for RoSTa integration is to employ ensemble by taking the median value of alpha mattes obtained by the models of all RoSTa, denoted as \textit{EN-median} in Table~\ref{tab:ensemble}. The results are better than each individual RoSTa on all evaluation metrics, $i.e.$ SAD 10.62 to 11.28, 11.76 and 12.26. However, the test speed of EN-median is quite slow comparing with others. To address this issue, we designed a new variant of GFM with a RoSTa Integration Module (RIM) to make use of all three representations in a learnable manner. Specifically, we modified the last layer of GD and FD to three RoSTa-specific layers, where each pair of outputs from a specific RoSTa layer goes through a Collaboration Matting(CM) module to obtain the alpha matte. The three alpha mattes are then concatenated together and fed into RIM. RIM consists of a concatenation layer, an 1$\times$1 convolutional layer to transform the feature channel from 3 to 16, and a Squeeze-and-Excitation (SE) attention module~\citep{hu2018squeeze} to help re-calibrate the features and select the most informative ones to predict a better alpha matte via another 1$\times$1 convolutional layer. We adopted the same training loss as the original GFM, except that there are three groups of losses corresponding to the outputs of three RoSTa-specific layers and a CM loss on the final alpha matte after RIM. In this way, the network is trained to learn a better RoSTa integration.

As shown in Table~\ref{tab:ensemble}, the result of GFM-RIM is better than each individual RoSTa on four evaluation metrics and it achieves the best performance on MSE, even outperforming EN-median, $i.e.$ 0.0025 to 0.0026. It is also worth noting that GFM-RIM has comparable test speed with each individual RoSTa, while runs much faster than EN-median. In summary, GFM-RIM makes a good trade-off between performance and speed. The study of more effective RoSTa integration techniques deserves more effort.

\begin{table}[htbp]
\caption{Grid search results of the down-sampling ratio during test in GFM on AM-2k.}
\begin{center}
\resizebox{\linewidth}{!}{
\begin{tabular}{cc|ccccc}
\hline
$d_1$ & $d_2$  & SAD & MSE & MAD & Grad. & Conn. \\
\hline
2 & 2  & 12.26 &0.0041  &0.0072 &\underline{9.18} & 11.59\\
\hline
2.2 & 2.2  & 11.49 & 0.0036 &0.0067 & 9.22&10.81 \\
2.4 & 2.4  & 11.66 &  0.0036& 0.0068& 9.34& 10.99\\
2.6 & 2.6  & \underline{10.77} & \textbf{ 0.0029}& \underline{0.0062}&9.93 & \underline{10.10}\\
2.8 & 2.8  & 10.96 &  0.0030& 0.0064& 10.06&10.28 \\
\hline
3 & 3  & 11.28 & 0.0030 & 0.0065&10.89&10.63 \\
4 & 4  &15.99  &0.0054  &0.0095 &14.75& 15.35\\
\hline
3 & 2  & \textbf{10.26} &\textbf{ 0.0029} & \textbf{0.0059}&\textbf{8.82}& \textbf{9.57}\\
4 & 2  & 12.96 &  0.0046& 0.0077&9.58&12.27 \\
4 & 3  & 13.97 &0.0048  & 0.0083&11.61&13.32 \\
\hline
\end{tabular}}
\end{center}
\label{tab:hybrid}
\end{table}

\subsubsection{Hybrid-resolution Test}

To investigate the influence of balancing GD and FD with different down-sampling ratios during the test, we conducted several experiments and reported the results in the Table~\ref{tab:hybrid}. For simplicity, we denote the down-sampling ratio at each step as $1/d_1$ and $1/d_2$, which are subject to $d_1\in\left \{ 2, 3, 4\right \}$, $d_2\in\left \{ 2, 3, 4\right \}$, and $d_1\geq d_2$. A larger $d_1$ increases the receptive field and benefits the GD, while a smaller $d_2$ benefits the FD with clear details in high-resolution images. We then find out that when using a hybrid-resolution test strategy, the performance achieves the best when $d_1=3$ and $d_2=2$, but with the cost of slower inference time.

\begin{table*}[htb]
\caption{Comparison of model parameters, computational complexity, and inference time. $(d)$, $(r)$ and $(r\dag)$ stand for DenseNet-121~\citep{huang2017densely} and ResNet-34, and ResNet-101~\citep{he2016deep}.}
\begin{center}
\begin{tabular}{l|ccc}
\hline
Method & Parameters (M) & Complexity (GMac) & Inference time (s)  \\
\hline
 SHM~\citep{chen2018semantic} & 79.27 &870.16 & 0.3346\\
 LF~\citep{zhang2019late} &37.91  & 2821.14& 0.3623\\
 HATT~\citep{Qiao_2020_CVPR} &106.96 &1502.46 & 0.5176\\
 SHMC~\citep{liu2020boosting} &78.23 &139.55 &0.4863 \\
 GFM-TT(d) & 46.96  & 244.0& 0.2085\\
 GFM-TT(r) & 55.29  & 132.28 & 0.1734\\
 GFM-TT(r2b) & 126.85  & 1526.79 &0.2268\\
 GFM-TT(r$^\dag$) & 497.26& 852.45& 0.1943\\
\hline
\end{tabular}
\end{center}
\label{Tab:run_time_parameters}
\end{table*}

To address this issue, we searched the optimal hyper-parameter setting of the down-sampling ratio to make a better balance between the two decoders. As can be observed from Table~\ref{tab:hybrid}, GFM achieves the best when the ratios are $1/2$ for FD and $1/3$ for GD. Accordingly, we tried to search for the best hyper-parameter that is the same for both decoders, $i.e.$, from $1/2$ to $1/3$ with a step of 0.2. As can be seen, GFM achieves the best performance when $d_1=d_2$ and equals to 2.6, $i.e.$, resizing the test image to $5/13$ of its original size. The results on all metrics are comparable to the previously best setting, $i.e.$, a hybrid-resolution ratio of $1/2$ and $1/3$. It is also noting that there is only a single forward pass when using the same ratio for both decoders, which is much more computationally efficient than the hybrid-resolution test.

\subsection{Model Complexity Analysis}
\label{sec: complexity}

We compared the number of model parameters (million, denoting ``M''), computational complexity (denoting ``GMac''), and inference time (seconds, denoting ``s'') of each method on an image resized to $800\times800$. All methods are performed on a server with an Intel Xeon CPU (2.30GHz) and an NIVDIA Tesla V100 GPU (16GB memory). As shown in Table~\ref{Tab:run_time_parameters}, GFM using either DenseNet-121~\citep{huang2017densely} or ResNet-34, or ResNet-101~\citep{he2016deep} as the backbone surpasses SHM~\citep{chen2018semantic}, LF~\citep{zhang2019late}, HATT~\citep{Qiao_2020_CVPR}, and SHMC~\citep{liu2020boosting} in running speed, $i.e.$, taking about 0.2085s and 0.1734s to process an image. In terms of parameters, GFM has fewer parameters than all the SOTA methods except for LF~\citep{zhang2019late}. For computational complexity, GFM has fewer computations than all the SOTA methods when adopting ResNet-34~\citep{he2016deep} as the backbone, $i.e.$, 132.28 GMacs. When adopting DenseNet-121~\citep{huang2017densely}, it only has more computations than SHMC~\citep{liu2020boosting} while being smaller. As for GFM(r2b) and GFM($r\dag$), they have more parameters and computations, but similar inference time. Although it can achieve better results, a trade-off between performance and complexity should be made for practical applications. Generally, GFM is light-weight and computationally efficient.

\section{Conclusion and Future Works}

In this paper, we propose a novel deep matting model for end-to-end natural image matting. It addresses two challenges in the matting task: 1) recognizing various foregrounds with diverse shapes, sizes, and textures from different categories; and 2) extracting details from ambiguous context background. Specifically, a Glance Decoder is devised for the first task and a Focus Decoder is devised for the latter one, while they share an encoder and are trained jointly. Therefore, they collaboratively accomplish the matting task and achieve superior performance than state-of-the-art matting methods. Besides, we also investigate the domain discrepancy issue between composite images and natural ones, which suggests that the common practice for data augmentation may not be suitable for training end-to-end matting models. To remedy this issue, we establish two large-scale real-world image matting datasets AM-2k and PM-10k, which contains 2,000 high-resolution animal images from 20 categories and 10,000 high-resolution portrait images along with the manually labeled alpha mattes. Furthermore, we systematically analyze the factors affecting composition quality including resolution, sharpness, semantic, and noise, and propose a novel composition route together with a large-scale background dataset BG-20k containing 20,000 high-resolution images without salient objects, which can effectively address the domain discrepancy issue. Extensive experiments validate the superiority of the proposed methods over state-of-the-art methods. We believe the proposed matting method and composition route will benefit the research for both trimap-based and end-to-end image matting. Moreover, the proposed datasets can provide a test bed to study the matting problem regarding the domain discrepancy issue.

Although GFM outperforms state-of-the-art methods in terms of both objective metrics and subjective evaluation, there are some limitations to be addressed in future work. First, after taking a detailed analysis of the error source as evidenced by SAD-TRAN, SAD-FG, and SAD-BG, the error in the transition areas is larger than SAD in the foreground and background areas, $i.e.$, 8.24 v.s. 2.01 or 7.80 v.s. 4.09, even if the size of transition areas is usually much smaller than that of foreground and background areas. It tells that the performance could be further enhanced by devising a more effective Focus Decoder as well as leveraging some structure-aware and perceptual losses. Second, there is still room to improve the composite-based models to match those trained using original images by domain adaptation methods~\citep{zhang2019category}, since the cost required to generate a composite dataset is much easier than constructing a natural images based one. Besides, given that alpha matting and alpha blending are inverse problems, it is interesting to see whether or not these two tasks benefit each other if we model them in a single framework. Third, current available natural image matting datasets contain only salient foreground images with small transition areas, e.g., human or animal, how to expand the dataset as well as extending the matting method to handle semi-transparent or long-range transition areas like a plastic bag or raindrops, remains as an open challenge. To this end, the two decoders in our GFM may be modified to adapt to different matting image types, $i.e.$, containing salient objects or semi-transparent objects. For example, a new representation like the trimap for the GD can be investigated to help the FD focus on explicit transition areas in those different types of images. Fourth, investigation of the interaction between the GD and FD to enable a more effective collaborative learning scheme also deserves more effort.

\bibliographystyle{spbasic}
\bibliography{2021_ijcv_matting.bib}

\end{document}